%% file: main.tex
\documentclass{article}

\usepackage[preprint]{corl_2026}

\usepackage{amsmath,amssymb,amsfonts,mathtools}
\usepackage{graphicx}
\usepackage{textcomp}
\usepackage{booktabs}
\usepackage{caption}
\usepackage{subcaption}
\usepackage{adjustbox}
\usepackage{multirow}
\usepackage{array}
\usepackage{enumitem}
\usepackage{wrapfig}
\usepackage[capitalise]{cleveref}

\setlist[itemize]{leftmargin=1.15em,itemsep=0pt,topsep=1pt,parsep=0pt,partopsep=0pt}
\setlist[enumerate]{leftmargin=1.3em,itemsep=0pt,topsep=1pt,parsep=0pt,partopsep=0pt}

\newcommand{\bb}{\mathbf}

\begin{document}
\title{TriPilot-FF: Guided Teleoperation for Bimanual Mobile Manipulation Policy Learning}
\author{
  \textbf{Zihao Li\textsuperscript{1}}, \textbf{Yanan Zhou\textsuperscript{2}},
  \textbf{Ranpeng Qiu\textsuperscript{3}}, \textbf{Hangyu Wu\textsuperscript{1}},
  \textbf{Guoqiang Ren\textsuperscript{1}}, \textbf{Weiming Zhi\textsuperscript{2}}\\
  $^{1}$Zhejiang University\\
  $^{2}$The University of Sydney\\
  $^{3}$Zhejiang University of Technology
}
\maketitle

\begin{abstract}
Mobile manipulators extend robot manipulation beyond the reach of fixed-base arms, but whole-body teleoperation remains difficult. Operators must coordinate a mobile base and two arms while reasoning about obstacles, contact, and arm reachability, often from incomplete visual feedback. We present \emph{TriPilot-FF}, an open-source feedback-guided teleoperation system for a custom bimanual mobile manipulator. TriPilot-FF uses a foot-operated pedal for continuous base control and augments it with lidar-driven force feedback: a low-cost base-mounted lidar estimates obstacle proximity in the commanded direction and renders a resistive cue at the pedal, discouraging unsafe base motion while preserving direct human control. The system combines this with bimanual leader-follower arm teleoperation, arm-side force reflection for contact awareness, and real-time manipulability guidance that prompts base repositioning when the arms approach poor configurations. Together, these feedback channels expose task-relevant constraints at the point of command selection, producing demonstrations that better reflect collision avoidance, contact regulation, and whole-body coordination. Across a series of real-world and long-horizon bimanual mobile manipulation tasks, TriPilot-FF acts as a lightweight co-pilot that improves operator coordination. We further show that feedback signals recorded during teleoperation can be incorporated into a visuomotor policy to improve autonomous execution.
\end{abstract}

\section{Introduction}

Mobile manipulators expand robot manipulation beyond the workspace of fixed-base arms. By moving the base, they can approach distant objects, recover from poor arm configurations, and complete long-horizon tasks without repeated re-grasping. However, this added capability also makes teleoperation substantially harder. The operator must coordinate base motion and bimanual manipulation while monitoring obstacle clearance, contact forces, and arm reachability, often through limited camera views and delayed visual consequences. This difficulty is not only an interface problem; it is also a data problem for robot learning \cite{intelligence2025pi_, zhu2026emma}. Modern visuomotor policies depend on demonstrations collected through human-operated systems \cite{RT_X}. If the interface hides important constraints, the recorded data may contain late braking, ambiguous recovery motions, excessive contact, or base repositioning only after the arms have entered poor kinematic configurations. A good teleoperation system should therefore do more than map human inputs to robot commands. It should help the operator perceive task-relevant constraints before actions are selected, so that the resulting demonstrations encode safer and more coordinated whole-body behavior. 

We present \emph{TriPilot-FF}, an open-source feedback-guided whole-body teleoperation system for a custom bimanual mobile manipulator. Its central idea is to render missing constraints directly through the teleoperation interface. Feet are usually overlooked in teleoperation setups. Here, TriPilot-FF uses a foot-operated pedal to control the mobile base, while lidar-driven force feedback renders obstacle proximity as a resistive cue. When the operator commands motion toward nearby obstacles, the pedal pushes back, discouraging collision-prone commands without taking control away from the user. TriPilot-FF combines this base interface with bimanual leader-follower arm teleoperation and arm-side force reflection, allowing the operator to feel contact interactions during manipulation. It also provides real-time force and visual guidance from bimanual manipulability, prompting the operator to reposition the base before the arms reach poor configurations. These feedback channels target three recurring failure modes in mobile bimanual manipulation: close-clearance base collisions, unstable or excessive contact, and reachability-driven dead ends.

\begin{figure}[t]
    \centering
    \includegraphics[width=\linewidth]{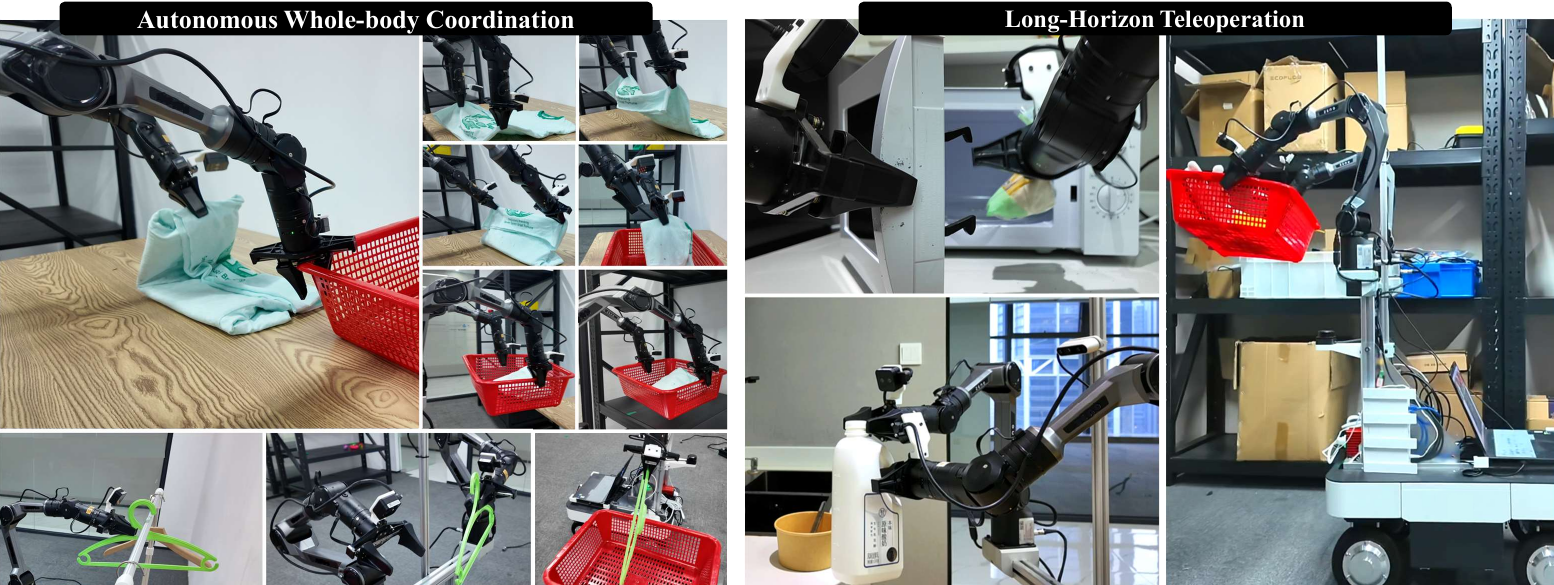}
    \caption{TriPilot-FF uses force feedback during teleoperation to shape demonstrations for mobile whole-body policy learning. The same sessions log images, proprioception, base commands, arm targets, motor efforts, and feedback signals for policy training and contact-rich analysis.}
    \label{fig:teaser}
    \vspace{-0.3em}
\end{figure}

Finally, TriPilot-FF connects feedback-guided teleoperation to autonomous policy learning. The system logs not only visual observations and whole-body commands, but also feedback signals that encode obstacle resistance, contact interaction, and kinematic difficulty. We incorporate these signals into a visuomotor policy \cite{Zhao2023LearningFB} and show that additional force information improves autonomous execution. \Cref{fig:teaser} illustrates representative tasks.

Our technical contributions are: (1) an open-source whole-body teleoperation system that combines foot-operated base control with bimanual leader-follower arm teleoperation;
(2) a lidar-driven pedal feedback mechanism that renders direction-conditioned resistive cues from obstacle proximity while preserving direct human control;
(3) a multimodal feedback interface with arm-side force reflection and manipulability-based guidance for contact awareness and base repositioning;
(4) an augmented visuomotor policy that uses teleoperation feedback to improve autonomous mobile manipulation.

\section{Related Work}

Teleoperation is a standard route for collecting robot-learning demonstrations. Systems such as RoboTurk, VR-based teaching, low-cost leader-arm platforms, OPEN TEACH, Open-TeleVision, UMI, and Mobile ALOHA enable useful visuomotor data collection across tabletop, dexterous, and mobile bimanual manipulation \cite{Mandlekar2018ROBOTURKAC,Zhang2017DeepIL,Gello,iyer2024openteach,cheng2024tv,chi2024universal,fu2024mobile, ding2025bunny}. These systems primarily improve command capture, embodiment accessibility, visual immersion, or deployment scale. TriPilot-FF addresses a complementary problem: during mobile bimanual teleoperation, operators often lack direct access to constraints that determine demonstration quality, including obstacle proximity, contact forces, and arm reachability. Rather than treating teleoperation as passive recording, TriPilot-FF renders these constraints back to the operator at action-selection time. This is related to bilateral teleoperation and force reflection \cite{hokayem2006bilateral}, but our goal is not a new stability or transparency theorem. We instead use bounded haptic cues for guided demonstration collection. RoboCopilot is also related in its human-in-the-loop view of imitation learning \cite{wu2025robocopilot}, but TriPilot-FF focuses on physical haptic guidance during mobile bimanual data collection rather than interactive policy correction. 

TriPilot-FF also builds on mobile manipulation and whole-body learning systems that coordinate base and arm motion through planning, control, teleoperation, or learned policies \cite{Planning_mobile,Mobile_MPC,TAMP_mobile,review_mobile,fu2024mobile,dass2024telemoma, jiang2025behavior}. TeleMoMa and Mobile ALOHA are strong references for mobile manipulation teleoperation and long-horizon mobile bimanual imitation \cite{honerkamp2025whole}.

\begin{wraptable}{l}{0.45\textwidth}
    \vspace{-0.2em}
    \centering
    \footnotesize
    \setlength{\tabcolsep}{3.2pt}
    \caption{Representative prior systems vs TriPilot-FF.}
    \label{related_systems}
    \begin{adjustbox}{max width=\linewidth}
    \begin{tabular}{lccc}
        \toprule
        Feature & TeleMoMa & RoboCopilot & TriPilot-FF \\
        \midrule
        Mobile-base command & \checkmark & \texttimes & \checkmark \\
        Foot-operated base control & \texttimes & \texttimes & \checkmark \\
        Bilateral arm feedback & \texttimes & \checkmark & \checkmark \\
        Obstacle-aware pedal haptics & \texttimes & \texttimes & \checkmark \\
        Manipulability-based guidance & \texttimes & \texttimes & \checkmark \\
        Additive base-intent fusion & \texttimes & \texttimes & \checkmark \\
        \bottomrule
    \end{tabular}
    \end{adjustbox}
    \vspace{-0.2em}
\end{wraptable}
TriPilot-FF enables teleoperation when the operator is not colocated, and improves the data-generation interface through lower-limb haptic base control, arm-side force reflection, manipulability guidance. \Cref{related_systems} summarizes these capability differences. For policy learning, we follow the standard chunked visuomotor imitation paradigm used in ACT-style policies and related architectures \cite{Zhao2023LearningFB,chi2023diffusionpolicy,levine2016end,liu2025factr}; the learning experiment evaluates the value of the collected feedback signals within a visuomotor policy.

\begin{figure}[t]
    \centering
    \includegraphics[width=\linewidth]{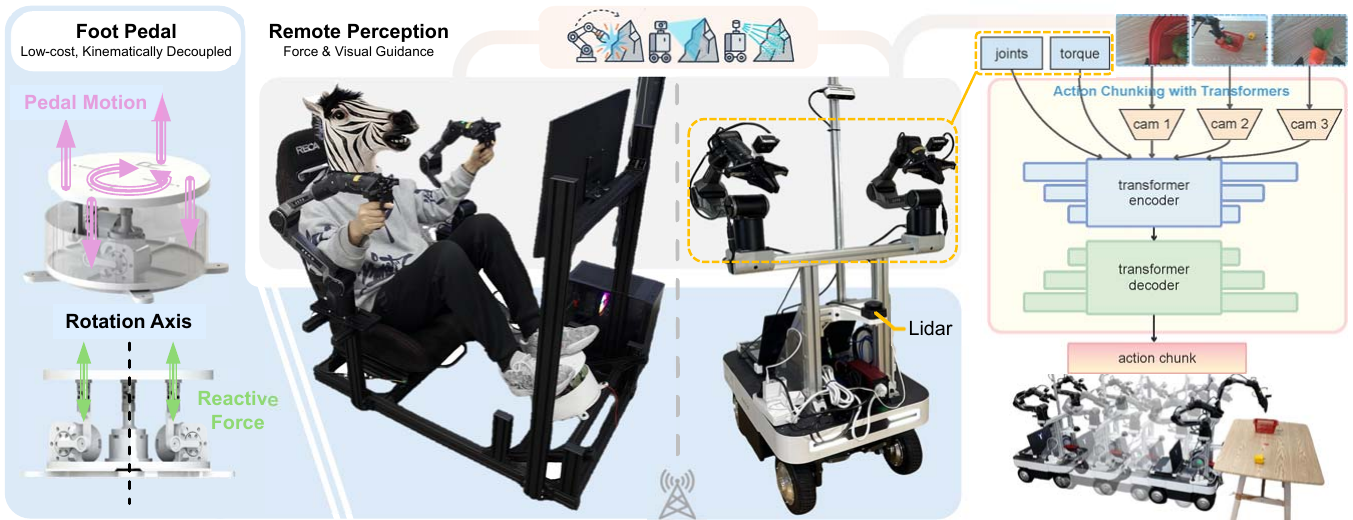}
    \caption{TriPilot-FF overview. The operator selects arm and base commands through leader arms and a motorized pedal. The robot-side lidar, follower efforts, gravity model, and reachability model produce bounded feedback or assistance during data collection. The resulting synchronized images, joints, efforts, base commands, and arm targets are then used for analysis and whole-body action-chunk imitation training.}
    \label{fig:system_diagram}
\end{figure}

\section{The TriPilot-FF Teleoperation System}
\label{sec:system}

\begin{wraptable}{r}{0.42\textwidth}
    \vspace{-1.em}
    \centering
    \scriptsize % smaller than \footnotesize
    % \fontsize{6.5pt}{7.2pt}\selectfont % use this instead of \scriptsize if you need even smaller text
    \setlength{\tabcolsep}{2.5pt}
    \renewcommand{\arraystretch}{1.02}
    \caption{TriPilot-FF feedback channels and their policy-learning role.}
    \label{tab:channels}
    \begin{adjustbox}{max width=\linewidth}
    \begin{tabular}{@{}
        >{\raggedright\arraybackslash}p{0.20\linewidth}
        >{\raggedright\arraybackslash}p{0.25\linewidth}
        >{\raggedright\arraybackslash}p{0.25\linewidth}
        >{\raggedright\arraybackslash}p{0.3\linewidth}
        @{}}
        \toprule
        Failure mode & Observed signal & Rendered cue & Effect \\
        \midrule
        Base clearance
        & Lidar point distances in the base frame
        & Thresholded pedal spring offset
        & Fewer low-clearance commands and abrupt reversals \\
        \midrule
        Contact instability
        & Filtered follower effort and gravity torque
        & Leader-arm feedforward torque
        & Smoother contact regulation and lower torque variation \\
        \midrule
        Low manipulability
        & Manipulability query and gradient
        & Additive base-intent velocity
        & Earlier base repositioning \\
        \bottomrule
    \end{tabular}
    \end{adjustbox}
    \vspace{-1em}
\end{wraptable}

TriPilot-FF is a whole-body teleoperation system for collecting high-quality demonstrations on a bimanual mobile manipulator. It assigns complementary control roles to the operator's hands and feet: the hands command the two arms through isomorphic leader-follower devices, while a 3-DoF foot pedal commands continuous planar base motion. The operator always retains authority, but TriPilot-FF renders task-relevant constraints before commands are logged. Lidar-based pedal resistance discourages collision-prone base motion, arm-side force reflection improves contact awareness, and manipulability guidance indicates when base repositioning would improve arm dexterity. \Cref{fig:system_diagram} summarizes the system and pedal design. \Cref{tab:channels} outlines the possible failures that occur during the teleoperation, and feedback cues that tackle each.

\subsection{System Hardware}
\label{subsect:hardware}

The teleoperation station synchronizes three information streams. First, the operator receives visual feedback from two wrist cameras, a head camera, and a low-cost 360-degree lidar. Second, bimanual leader arms provide arm and gripper commands. Third, a 3-DoF foot pedal provides base commands. The remote robot executes a unified whole-body command containing the planar base twist and dual-arm joint targets, while streaming proprioceptive and force-related signals for haptic feedback. We use AgileX Piper arms as both leader and follower devices.
\vspace{-0.3em}

\textbf{Pedal design.}
The pedal uses lower-limb dexterity for continuous base control, leaving the operator's hands free for manipulation. It senses and renders resistance along three approximately decoupled axes: $x$ translation, $y$ translation, and yaw. Three motors provide sensing and actuation, while the remaining components are 3D printed for low-cost replication. A turntable-linkage mechanism maps foot motion to the corresponding motor axis and reflects motor-generated resistance back along the same direction.
\vspace{-0.3em}

\textbf{Perception calibration.}
Constraint feedback requires camera observations, lidar measurements, end-effector poses, and manipulability estimates to share a common base-centered frame. We estimate the wrist-camera and head-camera extrinsics from multi-view checkerboard observations using an $SE(3)$ pose-graph formulation. For calibration pose $i$, let ${}^{B}\mathbf{T}_{G,i}$ be the base-to-gripper transform from forward kinematics, and let ${}^{C_h}\mathbf{T}_{\mathrm{tag},i}$ and ${}^{C_w}\mathbf{T}_{\mathrm{tag},i}$ be the tag poses observed by the head and wrist cameras. We solve for the fixed wrist-camera transform ${}^{G}\mathbf{T}_{C_w}$ and head-camera transform ${}^{B}\mathbf{T}_{C_h}$ such that$
{}^{B}\mathbf{T}_{G,i}
{}^{G}\mathbf{T}_{C_w}
{}^{C_w}\mathbf{T}_{\mathrm{tag},i}
\approx
{}^{B}\mathbf{T}_{C_h}
{}^{C_h}\mathbf{T}_{\mathrm{tag},i}$.
The aggregated residual is optimized on $SE(3)$ using a Lie-algebra parameterization and Levenberg--Marquardt.

\subsection{System Software}
\label{subsect:software}

At each control cycle, the operator provides leader-arm joint states and pedal commands. The logged whole-body action is $a_t=[u_{b,t}; q^L_t; q^R_t]$, where $u_{b,t}=(v_x,v_y,\omega)$ is the planar base twist and $q^L_t,q^R_t$ are the left and right follower-arm joint targets.
\vspace{-0.3em}

\textbf{Leader-follower arm feedback.}
Leader and follower arms run joint-space impedance control. The follower uses high-gain position tracking with targets copied from the leader, $q^{\mathrm{fol}}_{\mathrm{des}}(t)=q^{\mathrm{lead}}(t)$, and no feedforward torque. The leader uses lower gains so that it remains easy to move while still reflecting contact forces from the follower.

To reduce fatigue while preserving contact cues, we subtract model gravity from the follower torque and reflect only the estimated interaction component:
\begin{equation}
\tau^{\mathrm{lead}}_{i,\mathrm{ff}}
=
-\bigl(\tau^{\mathrm{fol}}_i-\hat g_i(q^{\mathrm{fol}})\bigr)s_i
+
\hat g_i(q^{\mathrm{lead}}),
\label{eq:force_reflection}
\end{equation}
where $s_i\in[0,1]$ controls the reflection strength for joint $i$. Here $\tau^{\mathrm{fol}}_i$ denotes the measured follower-side joint torque, $\hat g_i(\cdot)$ is the model-predicted gravity torque, and $\tau^{\mathrm{lead}}_{i,\mathrm{ff}}$ is the feedforward torque rendered on the leader joint.
\vspace{-0.3em}

\textbf{Collision-aware pedal resistance.}
Whole-body teleoperation often fails through small clearance errors while the operator is focused on manipulation. TriPilot-FF therefore renders pedal resistance against base commands that reduce obstacle clearance. Let $\hat d$ be the intended translational base direction from $(v_x,v_y)$, and let $r(\hat d)$ be the lidar-estimated free-space distance along that direction. A repulsive potential $\phi(r)$ produces a resistance magnitude $f_{\mathrm{rep}}(\hat d)=k_\phi |\partial \phi(r(\hat d))/\partial r|$. This resistance is applied at the pedal opposite to $\hat d$ and projected onto the available actuation axes. The feedback is therefore direction-conditioned: it discourages unsafe base motion without removing operator authority.
\vspace{-0.3em}

\textbf{Reachability guidance.}
Mobile manipulation often requires base repositioning before the arms enter low-dexterity configurations. We use manipulability as a dexterity proxy, $w(q)=\sqrt{\det(J(q)J(q)^\top)}$, and define a Cartesian manipulability field $m(x)$ as the best achievable manipulability at end-effector position $x$. Since solving this optimization online is impractical, we train a differentiable approximation $\hat m(x)$ from forward-kinematics samples.

The gradient $\nabla_x \hat m(x)$ gives a real-time direction for improving dexterity. We combine this direction with the current end-effector direction to form
\begin{equation}
v_{\mathrm{guide}}
=
\alpha \frac{x}{\|x\|}
+
(1-\alpha)
\frac{\nabla_x \hat m(x)}
{\|\nabla_x \hat m(x)\|}.
\label{eq:guide_direction}
\end{equation}
The horizontal component of $v_{\mathrm{guide}}$ is rendered through the pedal as a guidance force. Guidance is activated only when the arm is stretched and predicted manipulability is below a threshold. The same field can also be overlaid on camera views as a per-pixel reachability cue, helping operators identify comfortably reachable regions.

\subsection{Policy Learning}
\label{subsect:policy}

TriPilot-FF logs whole-body demonstrations for visuomotor imitation learning. We use ACT and augment its observation space with joint torques, which provide compact information about contact, interaction forces, and unmodeled dynamics. The policy predicts an action chunk as $\hat{\mathbf{a}}_{t:t+H}=\pi_\theta(\mathbf{q}_t,\boldsymbol{\tau}_t,\mathbf{I}_t)$, where $\mathbf{q}_t$ is proprioception, $\boldsymbol{\tau}_t$ is the measured torque signal, and $\mathbf{I}_t$ contains the camera observations. Each action contains base velocity and dual-arm joint targets, $\mathbf{a}_t=[\mathbf{v}^{\mathrm{base}}_t,\mathbf{q}^{\mathrm{arms}}_t]^\top\in\mathbb{R}^{2n+3}$, with $\mathbf{v}^{\mathrm{base}}_t=(v_x,v_y,\omega)$.

Torque inputs are embedded and injected into both the ACT encoder and decoder, so force information shapes the latent representation during training and conditions action generation during inference. The policy is trained with the standard ACT objective: an $\ell_1$ action-chunk prediction loss plus a KL regularizer on the latent variable. At deployment, overlapping predicted chunks are temporally aggregated with exponential weighting. Real-time torque measurements are fed back into the policy, enabling force-aware whole-body behavior.

\section{Experiments}
\label{sec:experiments}

\begin{figure}[t]
    \centering
    \begin{subfigure}[t]{0.41\linewidth}
        \centering
        \includegraphics[width=\linewidth]{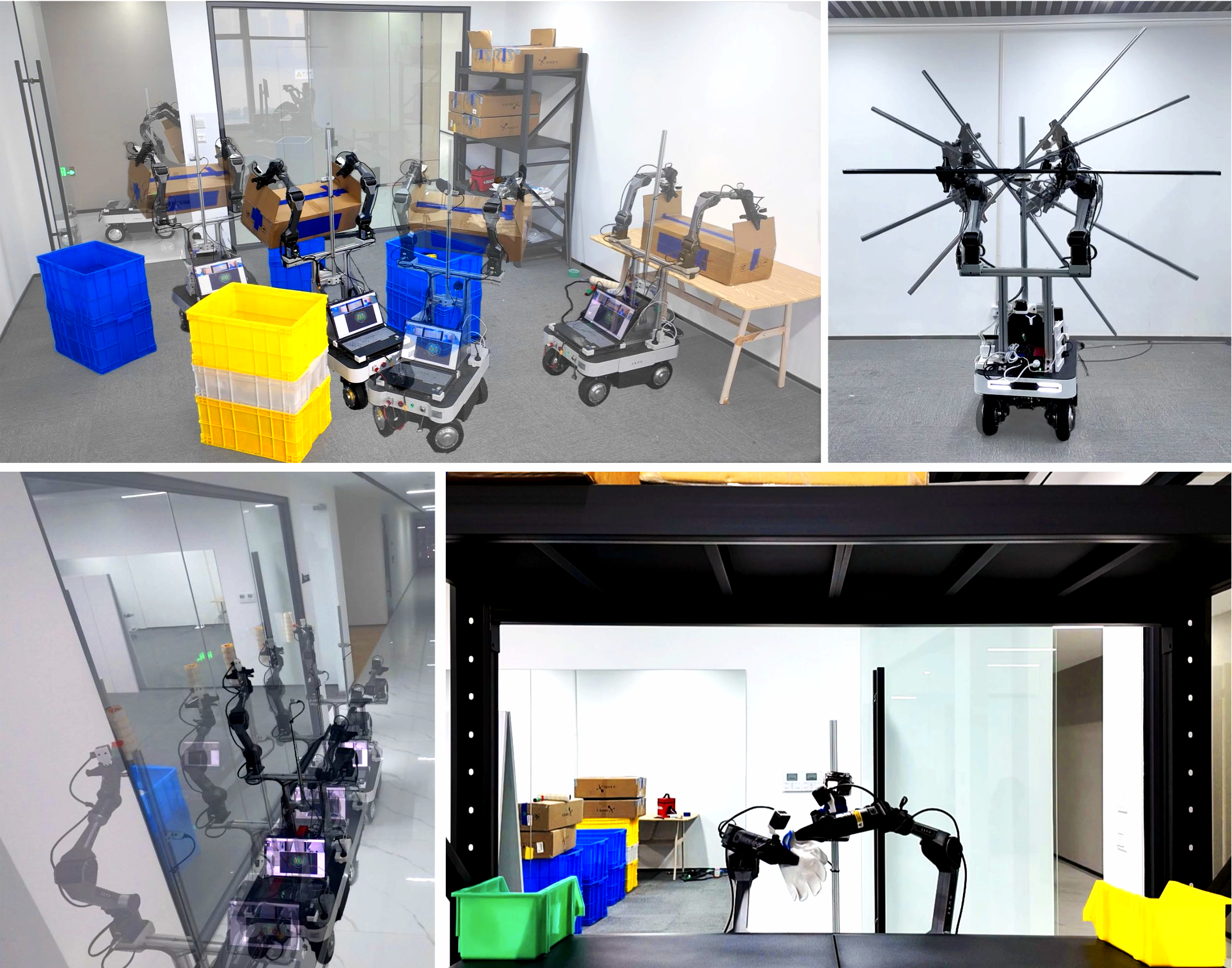}
        \caption{Task coverage used by the completed evaluation. The suite spans close-clearance mobile transport, long-object navigation, reachability-driven base repositioning, sustained contact, and long-horizon scenes.}
        \label{fig:task_suite}
    \end{subfigure}
    \hfill
    \begin{subfigure}[t]{0.57\linewidth}
        \centering
        \includegraphics[width=\linewidth]{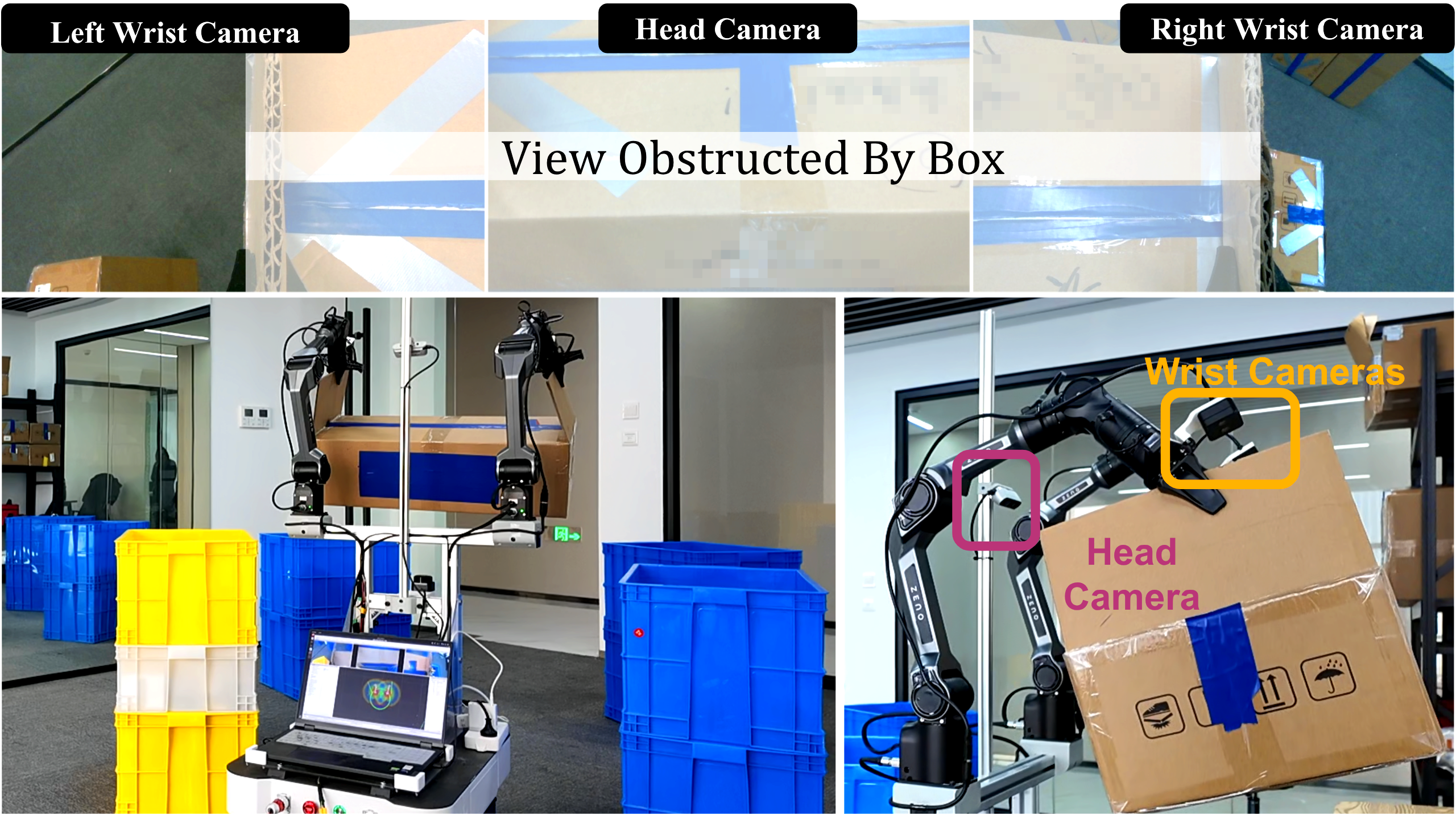}
        \caption{BlindCarry sensing problem. The carried box blocks the head-camera view of nearby obstacles, while the wrist cameras mainly see the grasp and object surface. The lidar-derived clearance signal is rendered at the pedal so collision risk is felt where the base command is selected.}
        \label{fig:blind_carry_1}
    \end{subfigure}
    \caption{\textbf{Task coverage and sensing challenge.} Left: the real-robot task suite used in the completed evaluation. Right: the BlindCarry obstruction setting, where local collision risk must be conveyed through pedal haptics because the carried object occludes the head-camera view.}
    \label{fig:task_suite_and_obstruction}
\end{figure}

We rigorously evaluate TriPilot-FF across a variety of tasks in the real-world task. We begin by evaluating TriPilot-FF on the following teleoperation tasks, with \Cref{fig:task_suite} showing an overview:
\vspace{-0.3em}
\begin{itemize}
    \item \textbf{BlindCarry} -- an occluded bimanual transport task where the robot carries a bulky object that blocks the forward camera view. The operator must maintain stable bimanual support while navigating clutter, stressing collision avoidance under poor visibility.

    \item \textbf{NarrowTransport} -- a constrained-passage transport task where the robot carries an elongated bar through a narrow doorway. Success requires coordinated base motion and bimanual posture adjustment to avoid boundary contact and maintain grasp alignment.

    \item \textbf{GuidedReach} -- a reach-limited retrieval task where the robot retrieves a glove from the left side of a shelf, hands it over between arms, and places it into a container on the right. The task tests whether manipulability guidance helps the operator reposition the base for better reachability.

    \item \textbf{MobileSwipe} -- a constrained contact task where the robot holds a roller against a vertical glass surface while the base moves laterally. The task tests whether arm-side force reflection improves contact regulation, reduces corrective motion, and supports stable whole-body contact.
\end{itemize}

\subsection{Feedback Guidance Improves Teleoperation}
\Cref{tab:teleop_results} summarizes the completed teleoperation ablations. \emph{BlindCarry} stresses the effect of pedal feedback for obstacle avoidance when camera occlusion occur, as illustrated in \cref{fig:blind_carry_1}. \emph{NarrowTransport} requires moving a long bar through a tight passage. \emph{GuidedReach} tests whether manipulability guidance encourages earlier base repositioning. \emph{MobileSwipe} tests sustained contact while the base moves laterally.

\begin{table}[t]
    \centering
    \scriptsize
    \setlength{\tabcolsep}{2.4pt}
    \caption{Completed haptic-guidance ablations. Values are aggregate readouts from the existing trials. Lower is better for average number of collisions, time, low-manipulability, torque variation, and energy.}
    \label{tab:teleop_results}
    \begin{adjustbox}{width=\linewidth}
    \begin{tabular}{lllll}
        \toprule
        Task & Metric & TriPilot-FF & Reduced feedback & Difference \\
        \midrule
        BlindCarry & Success / collisions / time & \textbf{100.0\%} / \textbf{0.00} / \textbf{131.34 s} & w/o pedal, 55.0\% / 1.00 / 150.08 s & +45.0 pp, -1.00 coll., -18.74 s \\
        \midrule
        NarrowTransport & Success / collisions / time & \textbf{100.0\%} / \textbf{0.00} / \textbf{16.32 s} & w/o pedal, 75.0\% / 0.30 / 48.36 s & +25.0 pp, -0.30 coll., -32.04 s \\
        NarrowTransport & Success / collisions / time & \textbf{100.0\%} / \textbf{0.00} / \textbf{16.32 s} & w/o any, 60.0\% / 0.40 / 49.16 s & +40.0 pp, -0.40 coll., -32.84 s \\
        \midrule
        GuidedReach & Low-manip. / time & \textbf{30.59\%} / \textbf{7.39 s} & w/o guidance, 41.04\% / 15.17 s & -10.45 pp, -7.78 s \\
        \midrule
        MobileSwipe & Torque std. / energy & \textbf{2.532} / \textbf{0.196} & w/o force, 3.239 / 0.313 & -0.707, -0.117 \\
        \bottomrule
    \end{tabular}
    \end{adjustbox}
\end{table}

The teleoperation ablations show that each feedback channel improves the failure mode it was designed to address. In \emph{BlindCarry} and \emph{NarrowTransport}, pedal resistance converts lidar-based clearance into an immediate command-space cue, helping operators avoid aggressive base motions near obstacles. This eliminates aggregate collisions in the reported trials and reduces completion time, especially when the forward camera is occluded or the carried object must pass through a narrow passage. In \emph{GuidedReach}, manipulability guidance reduces both task time and the fraction of low-dexterity states, indicating that operators reposition the base earlier instead of persisting in near-singular or over-extended arm configurations. In \emph{MobileSwipe}, arm-side force reflection reduces torque variation and energy use, consistent with smoother contact regulation during lateral base motion. Together, these results show that TriPilot-FF does more than make teleoperation subjectively easier: it improves success and efficiency, supports contact-rich whole-body behaviors such as wiping, and helps operators reach grasp targets across an extended workspace.

\begin{wraptable}{r}{0.45\textwidth}
    \centering
    \scriptsize
    \setlength{\tabcolsep}{3.2pt}
    \caption{Task-level NASA-TLX workload under task-specific ablations. Lower is better. $\Delta$ reports the increase relative to full TriPilot-FF.}
    \label{tab:nasa_tlx}
    \begin{adjustbox}{max width=\linewidth}
    \begin{tabular}{lcccc}
        \toprule
        Task & Full & Ablation & TLX & $\Delta$ \\
        \midrule
        \multirow{3}{*}{BlindCarry} 
        & \multirow{3}{*}{\textbf{30.7}} 
        & w/o pedal feedback & 58.7 & +28.0 \\
        & & w/o force feedback & 54.5 & +23.8 \\
        & & w/o any feedback & 59.6 & +28.9 \\
        \midrule
        \multirow{3}{*}{NarrowTransport} 
        & \multirow{3}{*}{\textbf{36.4}} 
        & w/o pedal feedback & 54.9 & +18.5 \\
        & & w/o force feedback & 58.1 & +21.7 \\
        & & w/o any feedback & 60.5 & +24.1 \\
        \midrule
        GuidedReach 
        & \textbf{33.2} 
        & w/o guidance & 42.6 & +9.4 \\
        \bottomrule
    \end{tabular}
    \end{adjustbox}
\end{wraptable}

\textbf{Operator workload study.}
We evaluate operator workload with 14 participants, including 11 novice and 3 experienced operators, across three tasks and 120 total trials. Participants completed a familiarization phase before performing the tasks with randomized condition order. Workload was measured using NASA-Task Load Index (TLX)~\cite{HART1988139}. As shown in \Cref{tab:nasa_tlx}, the full TriPilot-FF interface yields the lowest workload in every task. Across the study, the full interface also improves average success, reduces median completion time, and lowers collision events. The ablations are therefore best read as task-level removals from an integrated interface, not as perfectly isolated measurements of each feedback channel. Because the cues interact during teleoperation, removing even a secondary cue can increase workload, while removing the task-relevant cue, such as obstacle feedback in transport or reachability guidance in GuidedReach, produces the expected Task Load Index increase.

\subsection{Policy Learning with Torque-Augmented Whole-Body Policy}
\label{subsec:policy_torque_act}
We evaluate the quality of data collected by TriPilot-FF, and the impact of torque information in policy learning, both in simulation and in real-world deployment.

\vspace{0.2em}\noindent\textbf{Validation in Simulation:} In simulation, we evaluate two tasks, which we set up in MuJoCo \cite{todorov2012mujoco}: \textbf{CubeTransfer} requires grasping a randomly placed cube with one arm and transferring it to the other arm; \textbf{CubePickPlace} requires picking a cube from a randomly selected start pose and placing it at a goal location while coordinating mobile-base motion and dual-arm manipulation.
\begin{wraptable}{l}{0.35\linewidth}
  \centering
  \vspace{-0.5em}
  \caption{Results in MuJoCo, measured by success rate and reward}
  \label{tab:simulation}
  \setlength{\tabcolsep}{3.5pt}
  \begin{adjustbox}{max width=\linewidth}
  \begin{tabular}{l l r r}
    \toprule
    Task & Method & $S$ (\%) $\uparrow$ & $r_{\text{avg}}$ $\uparrow$ \\
    \midrule
    \multirow{2}{*}{CubeTransfer}
      & ACT                 & 22.0 & 826  \\
      & ACT + Torque        & $\mathbf{50.0}$ & $\mathbf{1222}$ \\
    \midrule
    \multirow{2}{*}{CubePickPlace}
      & ACT               & 28.0 & 1685 \\
      & ACT + Torque      & $\mathbf{36.0}$ & $\mathbf{1714}$ \\
    \bottomrule
  \end{tabular}
  \end{adjustbox}
  \vspace{-0.5em}
\end{wraptable}
We deploy the same whole-body control and logging stack in simulation, collecting 100 trajectories for training, and evaluating 50 rollouts of ACT policies. To keep supervision consistent across tasks, we use a staged progress reward, where the instantaneous reward at each stage is $\{0,1,2,3,4\}$. Each stage corresponds to a discrete milestone along the task trajectory. We report two evaluation metrics: (i) success rate $S$, defined as the fraction of episodes that reach the terminal stage within a fixed horizon, and (ii) average cumulative reward $r$ over all runs. Table~\ref{tab:simulation} summarises the results. Across both tasks, augmenting ACT with joint-torque signals improves performance, suggesting that torque provides task-relevant information that is difficult to infer from kinematics alone. In \emph{CubeTransfer}, adding torque increases success and substantially raises reward, indicating more reliable completion and fewer mid-trajectory failures. In \emph{CubePickPlace}, adding torque on top of base-velocity inputs yields a consistent success gain, with a modest improvement in return, consistent with the task being more driven by navigation.

\begin{wraptable}{l}{0.35\linewidth}
  \centering
  \vspace{-0.85em}
  \caption{We train policies on real-world data collected with TriPilot-FF.}
  \label{tab: q5_policy}
  \setlength{\tabcolsep}{3.5pt}
  \footnotesize
  \begin{adjustbox}{max width=\linewidth}
  \begin{tabular}{l l r}
    \toprule
    Task & Method & $S$ (\%) $\uparrow$ \\
    \midrule
    \multirow{5}{*}{BasketPack} & ACT & $12$ \\
                                     & ACT + Torque & $24$ \\
                                     & ACT + Co-training & $36$ \\
                                     % & ACT + Torque & \multirow{2}{*}{$\bb{60}$} \\
                                     % & $\qquad$ + Co-training &  \\
                                     & ACT + Torque &  \\
                                     & $\qquad$ + Co-training & $\bb{60}$ \\
    \midrule
    \multirow{2}{*}{HangerHandOff} & ACT & $68$ \\
                                         & ACT + Torque & $\bb{92}$ \\
                                         % & ACT + Co-training & -- \\
                                         % & ACT + Torque + Co-training & -- \\
    \midrule
    OOD- & ACT & $56$ \\
    HangerHandOff & ACT + Torque & $\bb{72}$ \\
    \bottomrule
  \end{tabular}
  \end{adjustbox}

\end{wraptable}

\begin{figure}[t]
    \centering

    \begin{subfigure}[t]{0.48\linewidth}
        \centering
        \includegraphics[width=\linewidth]{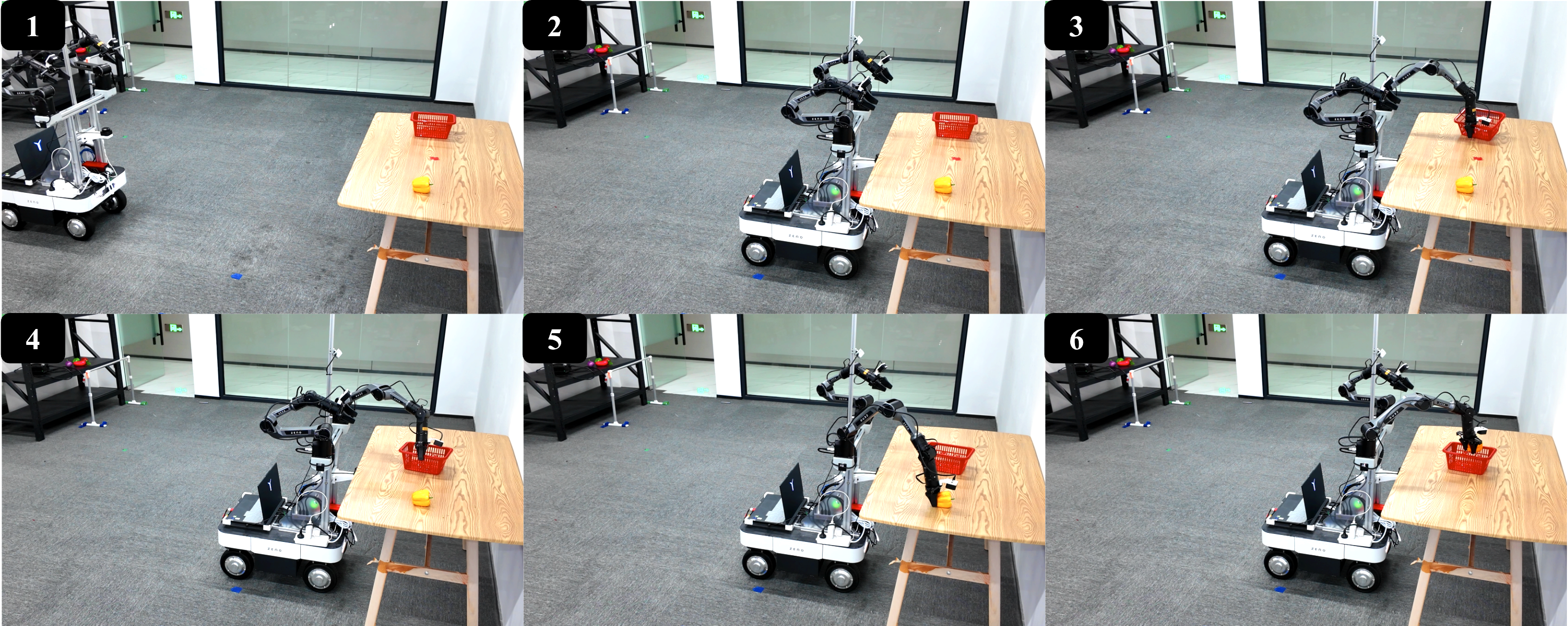}
        \caption{\textbf{BasketPlace.} The policy navigates to the table, centers the basket, and places the yellow pepper in.}
        \label{fig:basket_place}
    \end{subfigure}
    \hfill
    \begin{subfigure}[t]{0.48\linewidth}
        \centering
        \includegraphics[width=\linewidth]{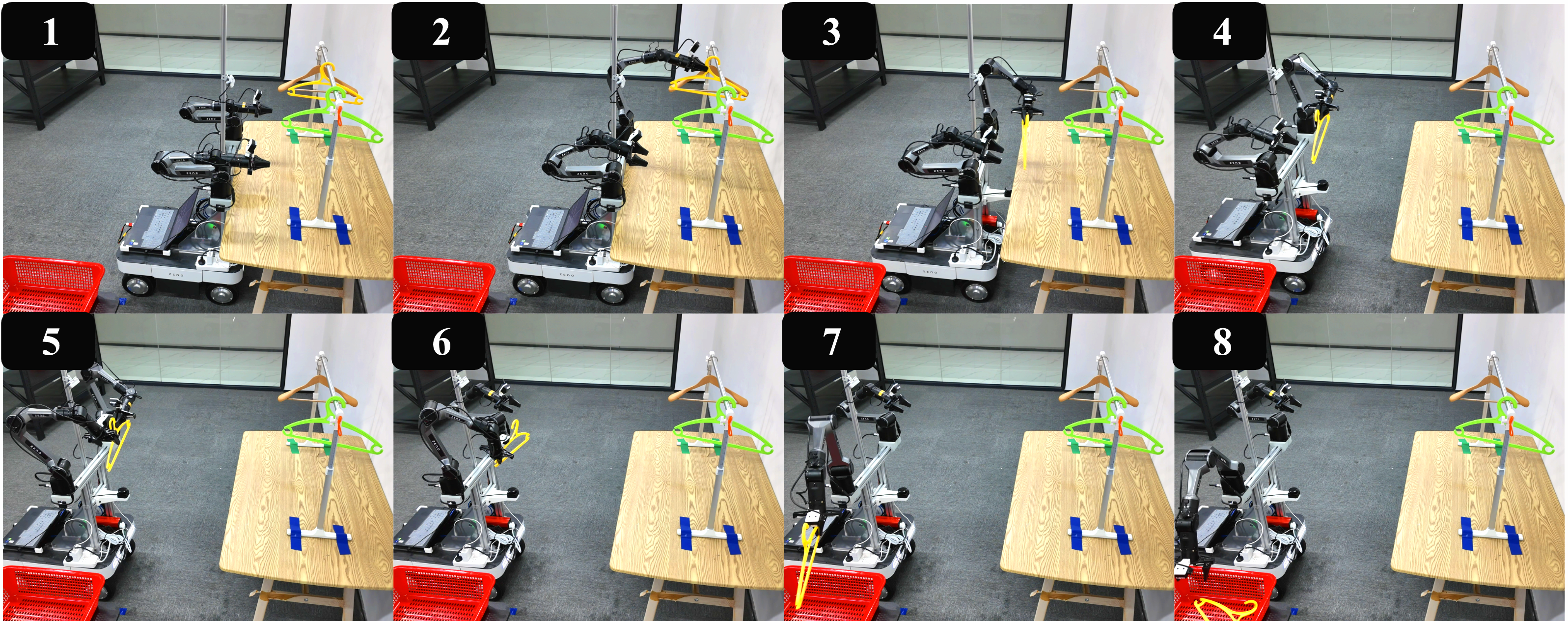}
        \caption{\textbf{OOD hanger transfer.} The policy handles unseen rack clutter, removes a hanger, transfers it between hands, moves to the basket and drops it in.}
        \label{fig:transfer}
    \end{subfigure}

    \caption{\textbf{Autonomous rollouts.} Policies trained from TriPilot-FF demonstrations execute mobile bimanual tasks requiring navigation, object transfer, and coordinated base--arm motion.}
    \label{fig:autonomous_rollouts}
\end{figure}

\noindent\textbf{Real-World Validation:} We evaluate policies learned on two constructed real-world bimanual mobile-manipulation tasks: \textbf{BasketPack} (\cref{fig:basket_place}), where a robot starts at a random distance away from a table, and needs to approach and stop at the table, then move a basket to the center, and move a toy pepper vegetable into the basket; \textbf{HangerHandOff} (\cref{fig:transfer}), where the robot removes a hanger from a clothes rack, moves while simultaneously performing a self-handover between its hands, and stows the hanger in a bin. For \emph{BasketPack}, we collect 20 trajectories where of whole-body robot motion. Then, similar to \cite{fu2024mobile}, we collect 100 trajectories of the robot stationary at the table with only upper-body motion, and add them to the dataset as \emph{co-training} data. For \emph{BasketPack}, as the motion of the upper limbs is highly coupled with the mobile base, we collect 100 trajectories of whole-body robot motion. We rollout 25 episodes per policy, and compute success rates, $S$, of completing the tasks.

\begin{wrapfigure}{r}{0.45\textwidth}
    \vspace{-0.5em}
    \centering
    \includegraphics[width=\linewidth]{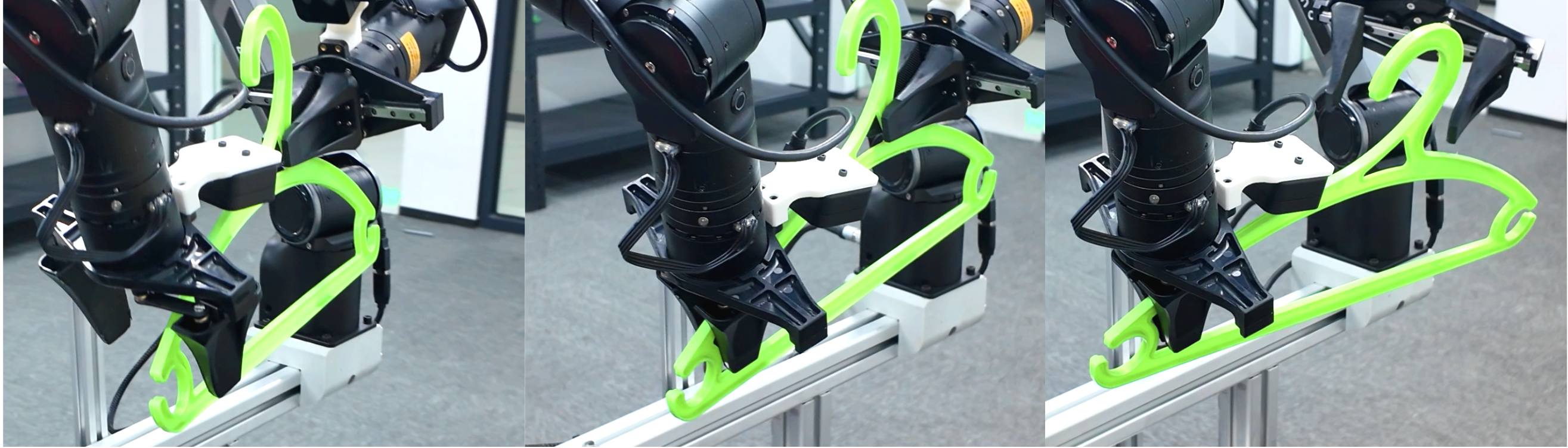}
    \caption{\textbf{In-motion transfer.} The torque-augmented policy produces smoother, more coordinated hanger transfer while the base is moving.}
    \label{fig:transfer_close}
    \vspace{-0.5em}
\end{wrapfigure}
To further test the ability for our model to adapt to unexpected events, we evaluate an \textbf{out-of-distribution (OOD)} variant of \emph{HangerHandOff}, where additional obstruction hangers, not observed in training, are placed on the rack. The performance of the trained ACT policies with and without force information is presented in \cref{tab: q5_policy}. We observe that, for both tasks, injecting the torque information into the state of the ACT policy provides significant performance increases. Additionally, an ACT policy for the \emph{BasketPack} task, trained with 20 trajectories, has a relatively low performance due to the limited amount of data. However, after integrating the additional upper-body motion data to \emph{co-train} the policy the performance improves significantly. Furthermore, we observe that the trained policy can handle the OOD variant with only a modest degradation in performance. Completing the \emph{HangerHandOff} task requires careful coordination between the two manipulators. This behaviour is shown in \cref{fig:transfer_close}.

\begin{figure}[t]
    \centering

    \begin{subfigure}[t]{0.5\linewidth}
        \centering
        \includegraphics[width=\linewidth]{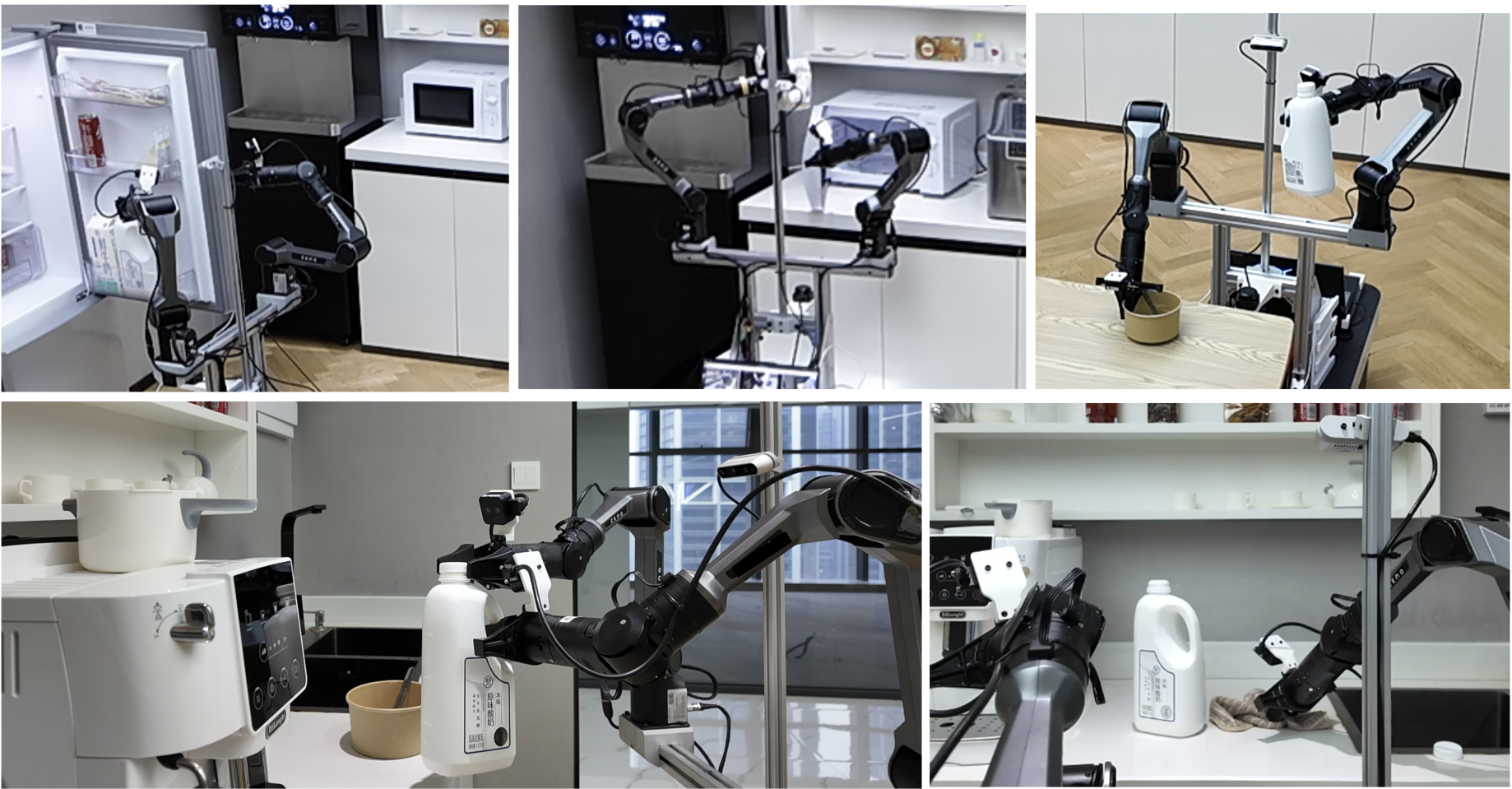}
        \caption{\textbf{Kitchen.} Long-horizon with fridge opening, microwave use, milk pouring, and bench wiping.}
        \label{fig:kitchen}
    \end{subfigure}
    \hfill
    \begin{subfigure}[t]{0.48\linewidth}
        \centering
        \includegraphics[width=\linewidth]{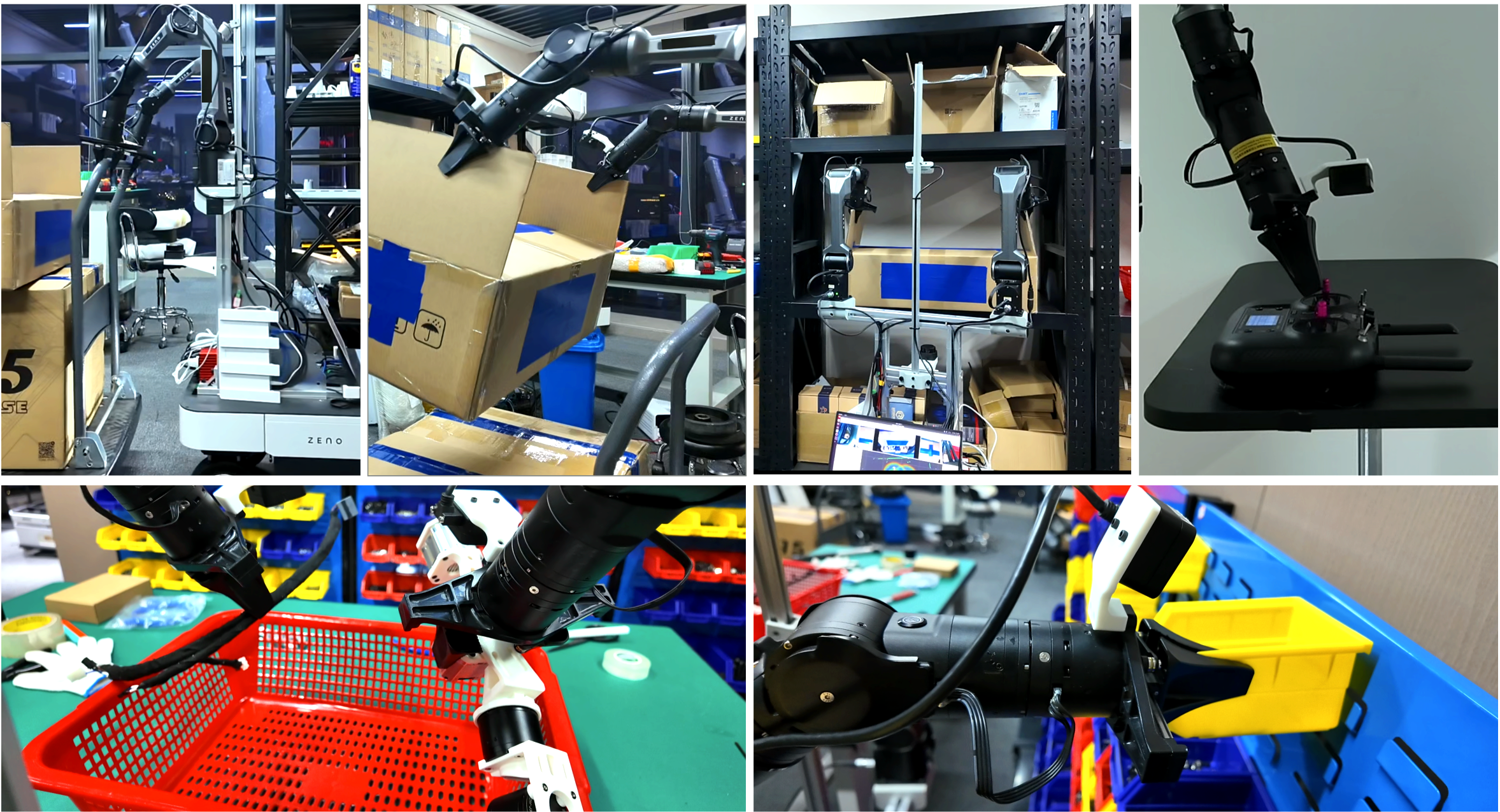}
        \caption{\textbf{Tool room.} Long-horizon with trolley pushing, box handling, shelf logistics, and bimanual packing.}
        \label{fig:tool}
    \end{subfigure}
    \caption{\textbf{Long-horizon teleoperation scenes.} TriPilot-FF supports extended mobile bimanual operation, requiring navigation, contact, object transfer, and coordinated base--arm motion.}
    \label{fig:long_horizon_scenes}
    \vspace{-0.5em}
\end{figure}

\vspace{-0.5em}
\subsection{Long-Duration Teleoperation and Autonomous Deployment}
\label{subsec:long_horizon}

\begin{wraptable}{r}{0.35\linewidth}
\vspace{-0.5em}
    \centering
    \setlength{\tabcolsep}{4.0pt}
    \caption{Repeated long-horizon teleoperation. Full success requires completing the whole scene without reset.}
    \label{tab:long_horizon}
    \begin{adjustbox}{max width=\linewidth}
    \begin{tabular}{lccc}
        \toprule
        Scene & Trials & Full success &Failures \\
        \midrule
        Kitchen & 5 & 80.0\% & 1 \\
        Tool room & 5 & 100.0\% & 0 \\
        \bottomrule
    \end{tabular}
    \end{adjustbox}
\end{wraptable}
While short-horizon benchmarks isolate specific failure modes (collision avoidance, reachability, and contact regulation), real deployments require \emph{task chaining}: the operator must sustain performance over extended episodes while switching between navigation, bimanual manipulation, and contact-rich interactions. To evaluate whether TriPilot-FF supports extremely long-duration teleoperation, we conduct long-horizon teleoperation in two representative scenes: a \textbf{kitchen} (\cref{fig:kitchen}) and a \textbf{tool room} (\cref{fig:tool}). In each scene, the operator completes \emph{multiple tasks in one continuous run} without any resetting. Each teleoperation sequence takes around 10 minutes. We provide videos of the robots operating in the supplementary materials. We repeat the collection of each sequence 5 times (\cref{tab:long_horizon}), and observe that TriPilot-FF enables reliable long-horizon teleoperation, with only a single trial failing.

\begin{wrapfigure}{r}{0.35\textwidth}
\centering
\vspace{-0.75em}
\includegraphics[width=\linewidth]{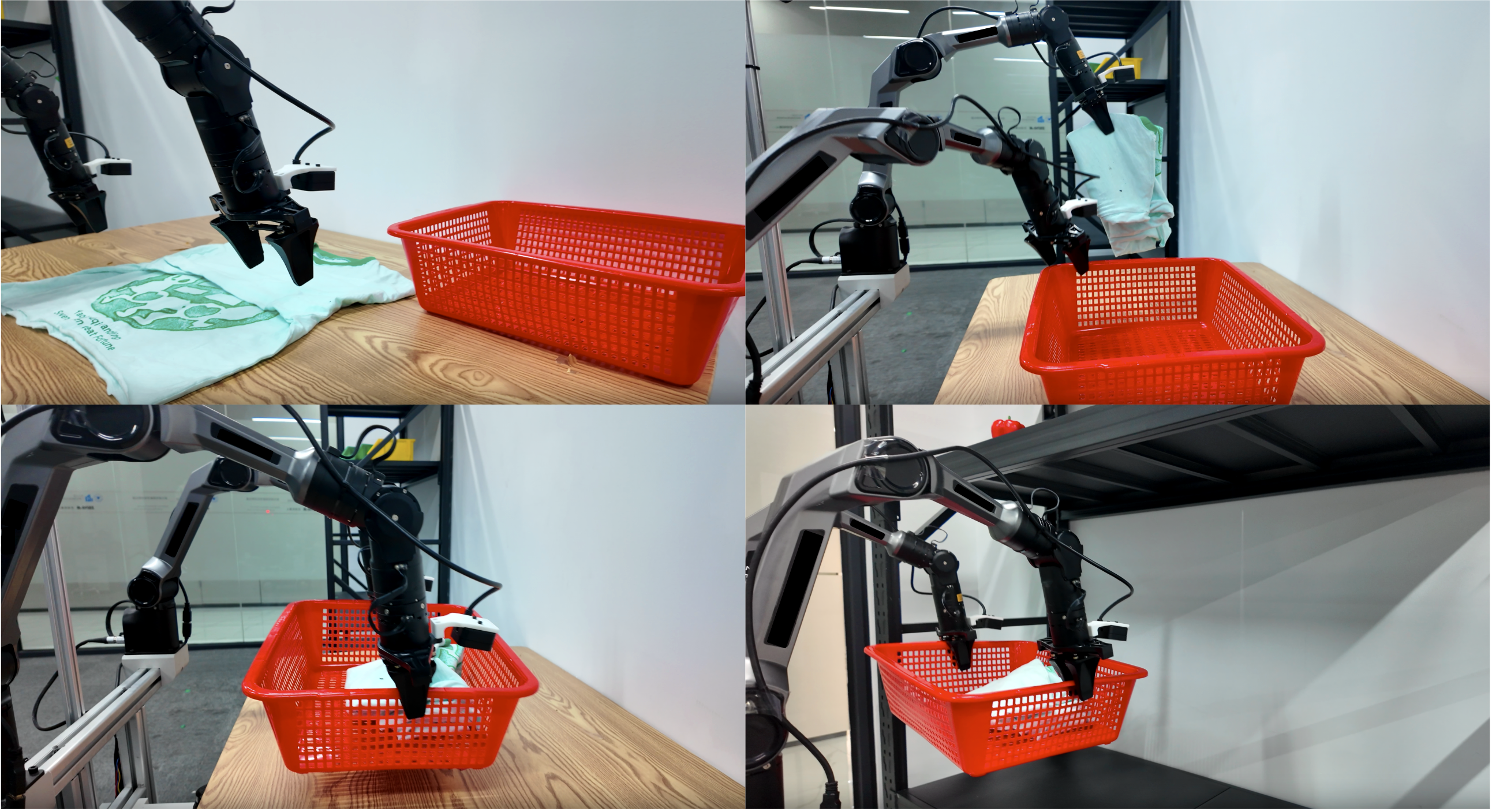}
\caption{Robot folds a T-shirt, packs it into a basket, and carries it to a shelf.}\label{Fig:long_hori}
\vspace{-0.75em}
\end{wrapfigure}

\noindent\textbf{Long-duration Autonomous Execution:} We additionally stress training autonomous robot behaviour with TriPilot-FF, by training a torque-augmented ACT model on a long-horizon task, \textbf{LaundryTransport}. Here, the robot needs to fold a t-shirt, place it into a basket, and then transport the basket to a shelf, taking around $1.5$ minutes. \Cref{Fig:long_hori} illustrates a successful execution of the task, with an additional deployment video in the supplementary.

\vspace{-1em}
\section{Conclusion and Limitations}
\vspace{-1em}
We present TriPilot-FF, a whole-body teleoperation system for bimanual mobile manipulators that combines foot-operated base control with bimanual leader-follower arm teleoperation and force feedback. TriPilot-FF renders resistive pedal cues that improve collision-averse driving in tight or occluded settings, and provides guidance on how to position the mobile base for greater reachability, while arm-side force reflection improves contact regulation in whole-body manipulation. We further show that teleoperation feedback signals are useful for learning: injecting joint-torque observations into a whole-body ACT policy improves imitation performance.
\vspace{-0.3em}

\textbf{Limitations:} 1. Although the guidance provided by TriPilot-FF enables operators to more efficiently teleoperate a mobile manipulator, this is still slow relative to fixed-base manipulation; 2. The current feedback channels improve obstacle awareness, contact regulation, and reachability, but do not yet adapt online to operator style, task phase, or predicted user intent. This shall be an area for future research.

\clearpage
\bibliography{ref}

\clearpage
\appendix
\input{appendix}

\end{document}

%% file: appendix.tex
\section{System Details}
\label{app:system}
\raggedbottom
\setlength{\parskip}{0pt}
\setlength{\topsep}{1pt}
\setlength{\partopsep}{0pt}
\setlength{\parsep}{0pt}
\setlength{\itemsep}{0pt}
\setlength{\abovecaptionskip}{1pt}
\setlength{\belowcaptionskip}{1pt}
\captionsetup{skip=1pt}
\renewcommand{\arraystretch}{0.88}
\begin{figure}[htbp]
    \centering
    \includegraphics[width=0.41\linewidth]{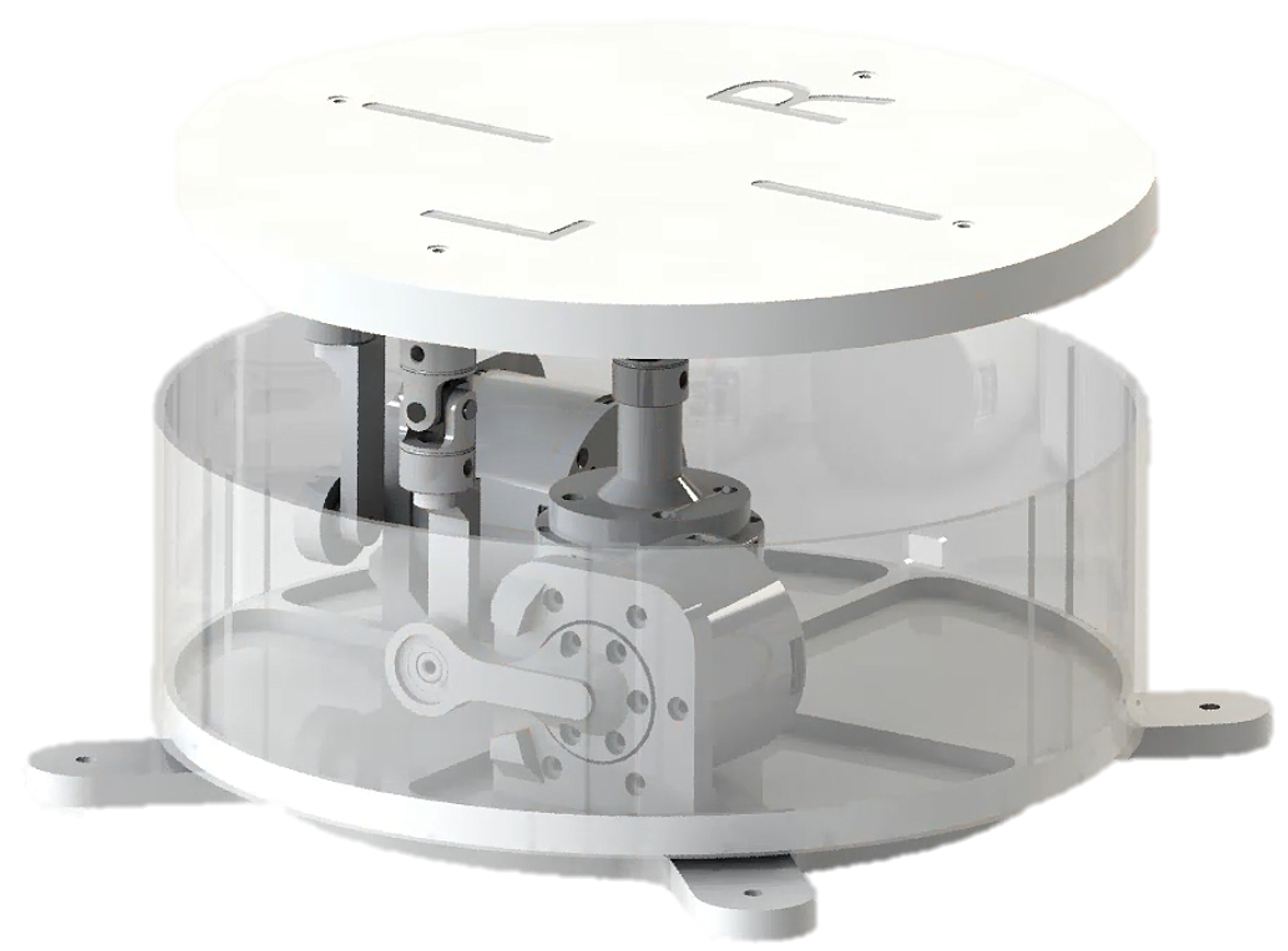}
    \includegraphics[width=0.56\linewidth]{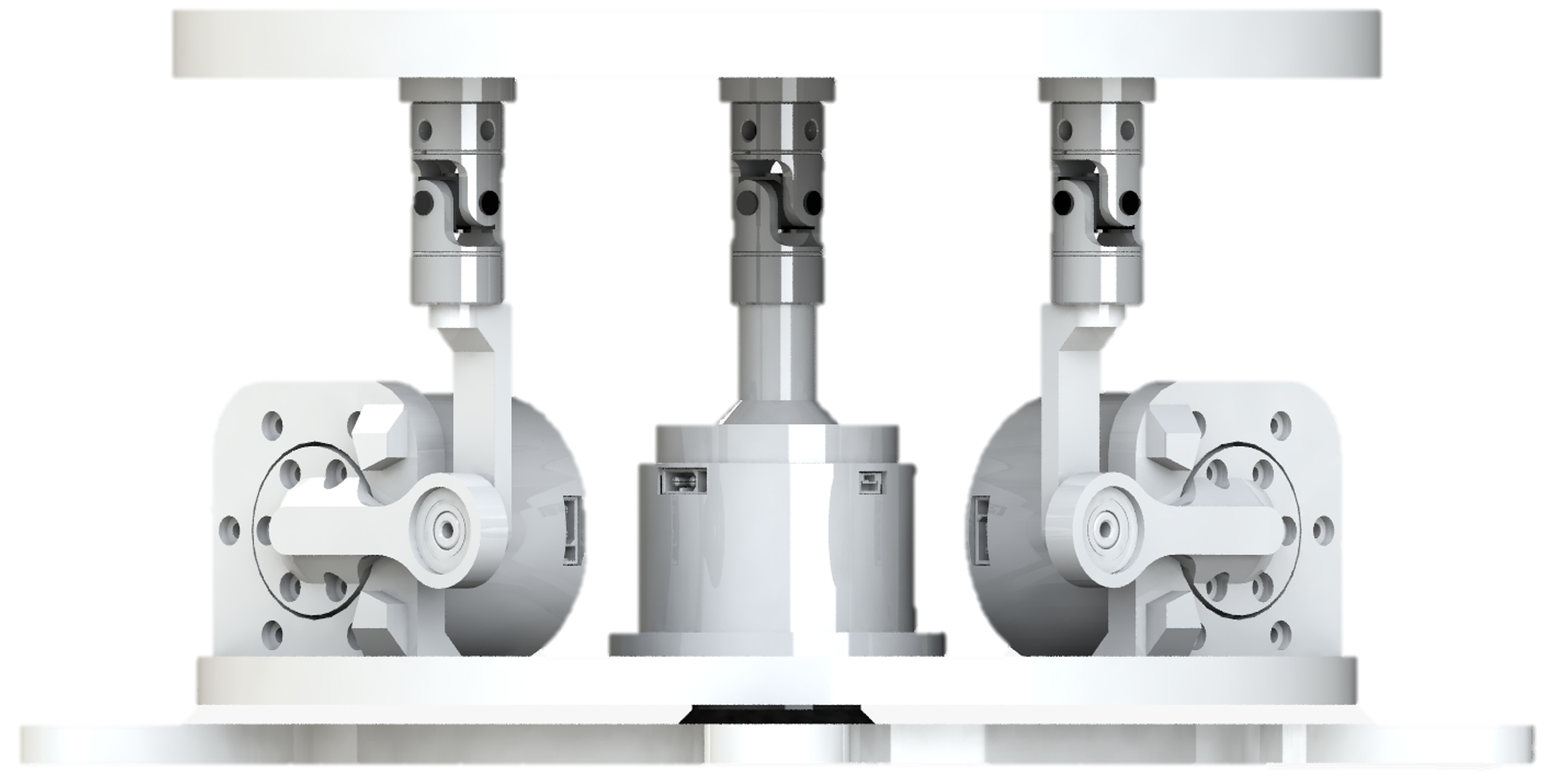}
    \caption{Motorized foot-pedal mechanism. The top plate is supported by three linkage-driven axes, allowing the device to sense foot displacement and render bounded resistance for forward, backward, lateral, and yaw base commands. The transparent CAD view exposes the compact 3D-printed housing, universal joints, and motor placement.}
    \label{fig:app_pedal}
\end{figure}

\begin{table}[htbp]
    \centering
    \footnotesize
    \caption{Material list for the teleoperation station and mobile bimanual robot.}
    \label{tab:app_material}
    \begin{adjustbox}{max width=\linewidth}
    \begin{tabular}{lll}
        \toprule
        Category & Component & Supplier / brand \\
        \midrule
        \multirow{5}{*}{Teleoperation station}
        & Structural frame & Lab assembled \\
        & Pedal structure & 3D printed \\
        & Pedal motors & dmBots DM-J4310 ($<\$50$ per motor) \\
        & Leader arms & AgileX Robotics PiPER \\
        & U-joints $\times3$ & Generic supplier \\
        \midrule
        \multirow{7}{*}{Mobile robot}
        & Structural frame & Lab assembled and 3D printed \\
        & Follower arms & AgileX Robotics PiPER \\
        & Power inverter & KLW \\
        & Wrist cameras & Intel RealSense D405 \\
        & Head camera & Intel RealSense D435 \\
        & Front lidar & SLAMTEC C1 Lidar \\
        & Mobile base & AgileX Robotics Ranger Mini V3 \\
        \bottomrule
    \end{tabular}
    \end{adjustbox}
\end{table}

\begin{table}[htbp]
    \centering
    \scriptsize
    \setlength{\tabcolsep}{3.0pt}
    \caption{Timing, synchronization, and logging summary. All streams use the ROS host clock, and policy tuples are aligned to command timestamps.}
    \label{tab:app_timing}
    \begin{adjustbox}{max width=\linewidth}
    \begin{tabular}{lcll}
        \toprule
        Stream & Rate & Alignment / latency & Use \\
        \midrule
        Lidar scan & 10 Hz & Newest scan near pedal command, $<10$ ms compute & Pedal resistance and clearance log \\
        Cameras & 30 Hz & Nearest earlier frame in a 50 ms window & Operator view and policy images \\
        Joint state & 100 Hz & Linear interpolation to policy/query time, $<5$ ms log path & Control and proprioception \\
        Torque / effort estimate & 100 Hz & Filtered motor effort and gravity-compensation topic & Arm reflection and contact/energy analysis \\
        Pedal sensing & 50 Hz & Newest displacement sample, 21.8 ms pedal-to-base command & Base action log \\
        Haptic rendering & 20 Hz pedal / 50 Hz arm & Stale inputs render zero cue & Feedback and cue log \\
        Manipulability network & 50 Hz & Query from joint state and base transform & Reachability overlay and base-intent command \\
        Policy inference & 10 Hz update / 50 Hz execution & Predicted chunks written to command buffer & Autonomous rollout \\
        \bottomrule
    \end{tabular}
    \end{adjustbox}
\end{table}

\section{Supplementary Evaluation Tables}
\label{app:extra_eval}

This section keeps quantitative material from earlier drafts that is useful for interpretation but not needed in the main evidence chain. The main text uses the code-supported whole-body action-chunk policy path as the primary policy comparison. The tables below preserve broader data-quality readouts for completeness without treating torque-conditioned policies or policy-family generality as implemented method components.

% \begin{table}[!htbp]
%     \centering
%     \footnotesize
%     \setlength{\tabcolsep}{3.4pt}
%     \caption{Representative prior systems and TriPilot-FF correspond to different teleoperation design points.}
%     \label{tab:app_related_systems}
%     \begin{adjustbox}{max width=\linewidth}
%     \begin{tabular}{lccc}
%         \toprule
%         Feature & TeleMoMa & RoboCopilot & TriPilot-FF \\
%         \midrule
%         Mobile-base command & \checkmark & \texttimes & \checkmark \\
%         Foot-operated base control & \texttimes & \texttimes & \checkmark \\
%         Bilateral arm feedback & \texttimes & \checkmark & \checkmark \\
%         Obstacle-aware pedal haptics & \texttimes & \texttimes & \checkmark \\
%         Manipulability-based guidance & \texttimes & \texttimes & \checkmark \\
%         Additive base-intent fusion & \texttimes & \texttimes & \checkmark \\
%         \bottomrule
%     \end{tabular}
%     \end{adjustbox}
% \end{table}

\begin{table}[!htbp]
    \centering
    \scriptsize
    \setlength{\tabcolsep}{3.0pt}
    \caption{Supplementary archival readouts on LaundryTransport/HangerHandOff. Each cell reports LaundryTransport / HangerHandOff. Bold marks the best or tied-best value within each task. DP denotes Diffusion Policy and is retained only as supplementary context, not as the deployed policy path emphasized in the main text.}
    \label{tab:data_quality}
    \begin{adjustbox}{max width=\linewidth}
    \begin{tabular}{lccc}
        \toprule
        Metric & Handheld & Teleop w/o FF & TriPilot-FF \\
        \midrule
        Collection rate (traj/hr) & 37.3 / 66.1 & 32.8 / 63.0 & \textbf{41.0 / 75.4} \\
        Unsafe demos (\%) & 3.0 / 1.0 & 17.0 / 4.0 & \textbf{0.0 / 0.0} \\
        Chunked policy success (\%) & \textbf{44.0} / 64.0 & 40.0 / 56.0 & \textbf{44.0} / \textbf{68.0} \\
        DP success (\%) & 40.0 / \textbf{60.0} & 32.0 / 56.0 & \textbf{44.0} / \textbf{60.0} \\
        \bottomrule
\end{tabular}
    \end{adjustbox}
\end{table}

These rows are not used as matched force-modality ablations or as evidence of policy-family generality. 

\section{Calibration And Retargetability}
\label{app:calibration}

\begin{figure}[htbp]
    \centering
    \includegraphics[width=0.32\linewidth]{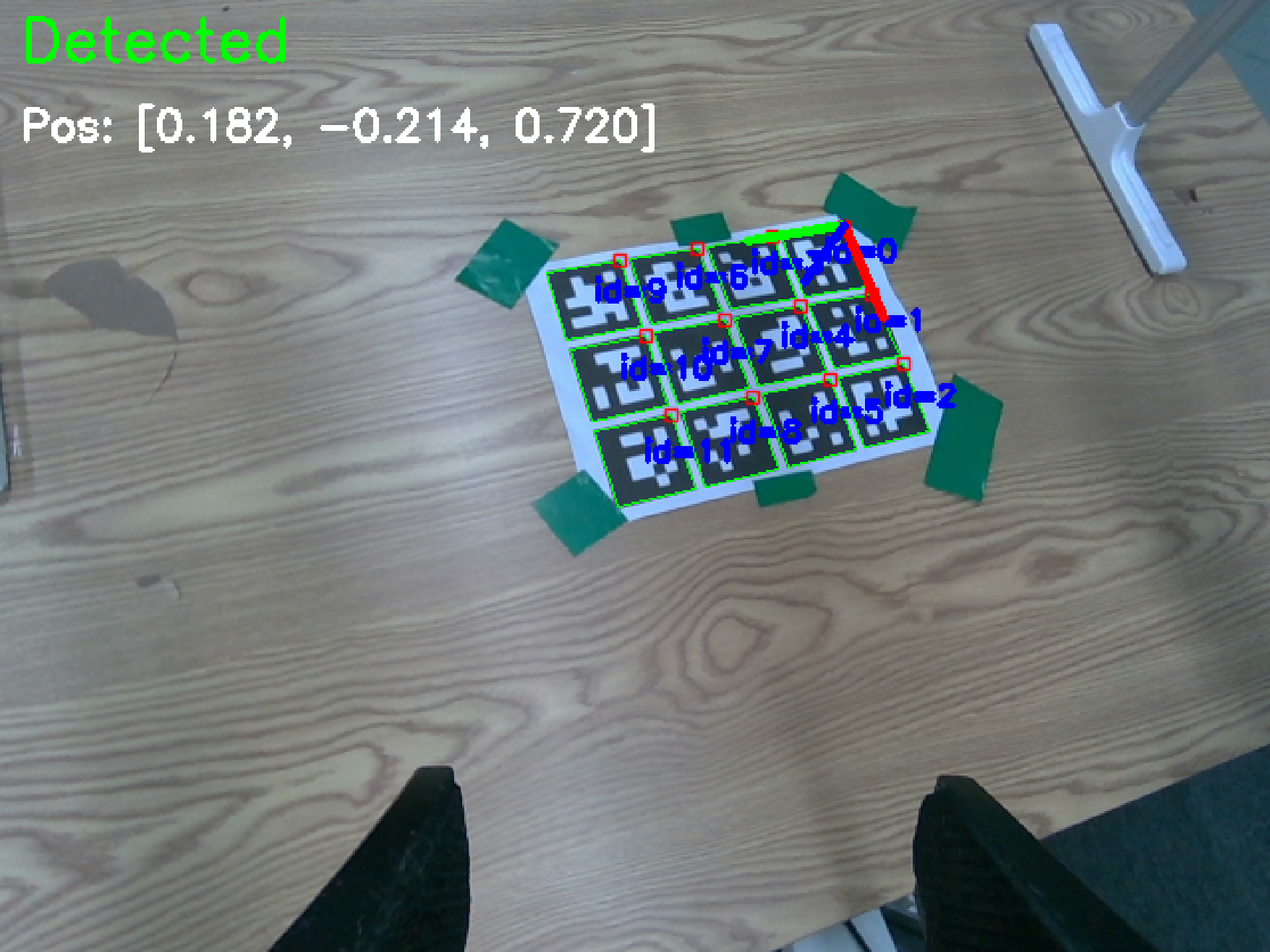}
    \includegraphics[width=0.32\linewidth]{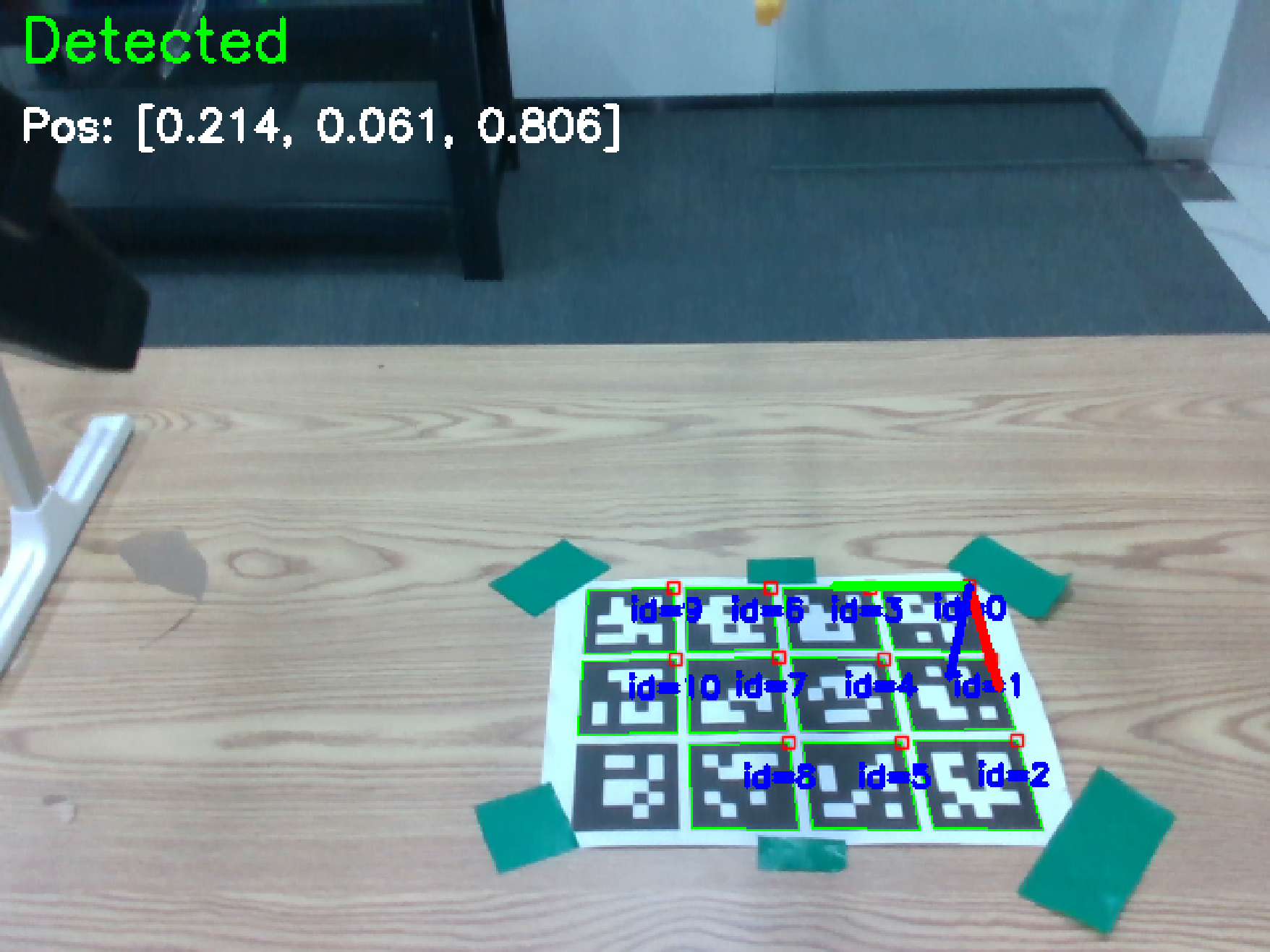}
    \includegraphics[width=0.32\linewidth]{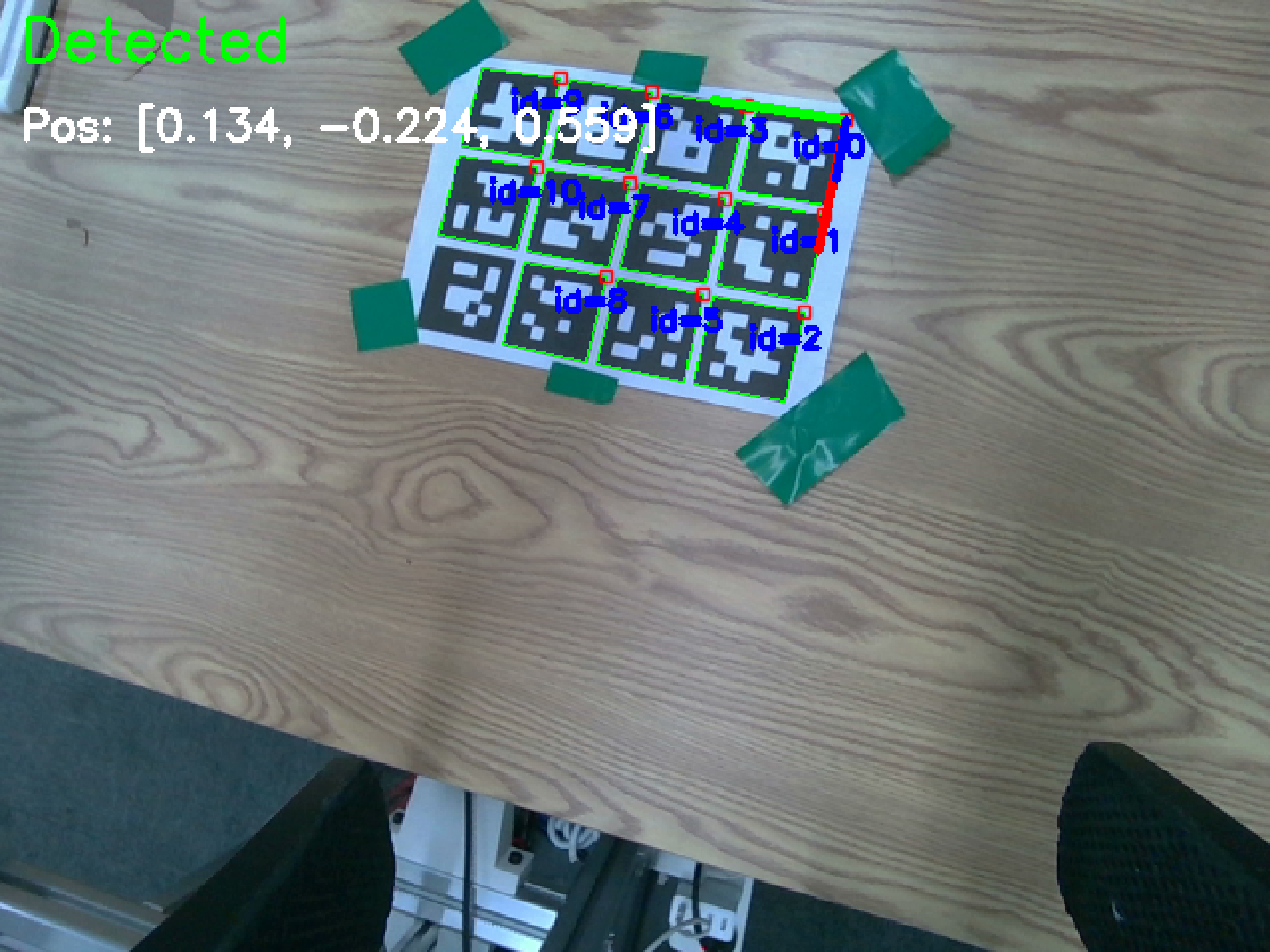}
    \caption{Calibration target observations from the left wrist, head, and right wrist cameras. The same fiducial board is observed from different robot poses so that wrist-camera extrinsics and the fixed head-camera transform can be solved in one base-centered frame.}
    \label{fig:app_calib_sample}
\end{figure}

TriPilot-FF overlays perception outputs and manipulability cues in a robot-centered coordinate system. Let ${}^{B}\mathbf{T}_{G,i}$ be the base-to-gripper transform from forward kinematics at configuration $i$. Let ${}^{C_h}\mathbf{T}_{\mathrm{tag},i}$ and ${}^{C_w}\mathbf{T}_{\mathrm{tag},i}$ be the fiducial target pose measured by the head and wrist cameras. The unknown transforms are ${}^{G}\mathbf{T}_{C_w}$ and ${}^{B}\mathbf{T}_{C_h}$. We solve
\begin{equation}
{}^{B}\mathbf{T}_{G,i}\,{}^{G}\mathbf{T}_{C_w}\,{}^{C_w}\mathbf{T}_{\mathrm{tag},i}
\approx
{}^{B}\mathbf{T}_{C_h}\,{}^{C_h}\mathbf{T}_{\mathrm{tag},i}
\end{equation}
by minimizing the aggregated $SE(3)$ residual over ten calibration configurations using Levenberg-Marquardt. Samples with low fiducial confidence or large reprojection error are discarded before optimization.

\begin{figure}[htbp]
    \centering
    \includegraphics[width=\linewidth]{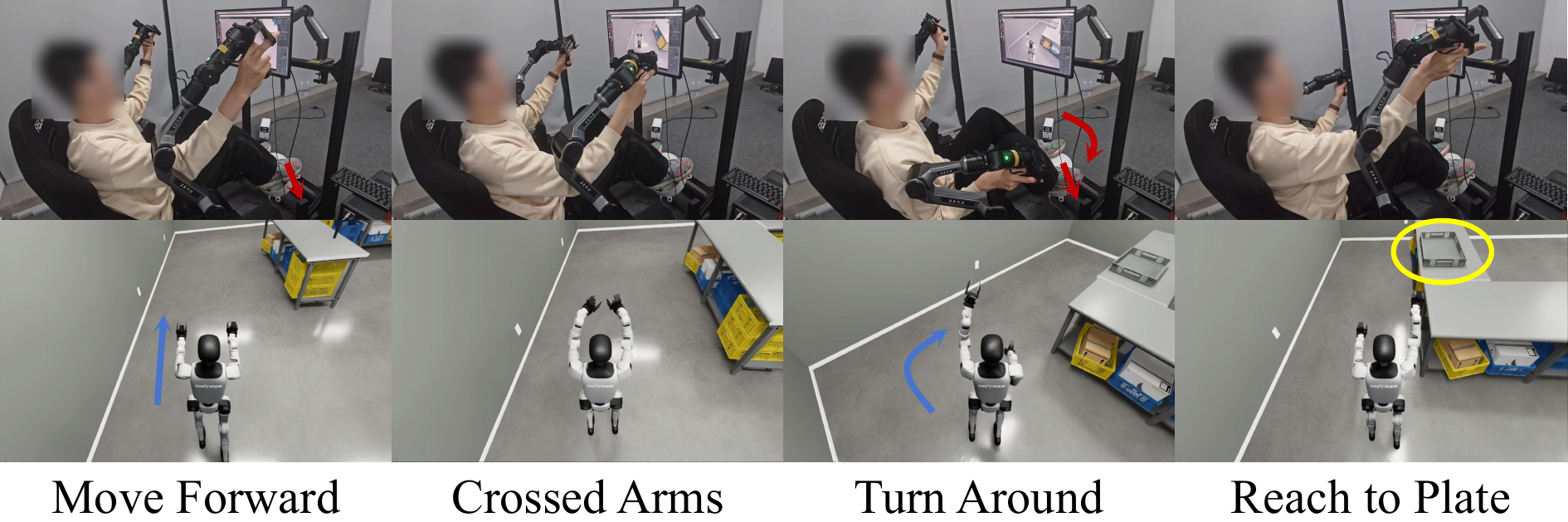}
    \caption{Interface retargetability example in simulation. The operator's bilateral leader-arm motion is replayed on a different humanoid-like embodiment through task-space inverse kinematics around matched home poses. The figure is included only to show that the leader interface can be retargeted. It is not evidence for cross-embodiment policy transfer.}
    \label{fig:app_cross_embodiment}
\end{figure}

\section{Policy And Control Hyperparameters}
\label{app:hyper}

\begin{table}[htbp]
    \centering
    \footnotesize
    \caption{Action-chunk policy hyperparameters used for the autonomous experiments.}
    \label{tab:app_act_hyper}
    \begin{adjustbox}{max width=\linewidth}
    \begin{tabular}{lll}
        \toprule
        Category & Hyperparameter & Value \\
        \midrule
        \multirow{6}{*}{Visual backbone}
        & Backbone & ResNet-18 \\
        & Pretrained weights & ImageNet1K\_V1 \\
        & Input image size & Dataset RGB frame, no ROS-side resize \\
        & Number of cameras & 3 \\
        & Image normalization mean & [0.485, 0.456, 0.406] \\
        & Image normalization std. & [0.229, 0.224, 0.225] \\
        \midrule
        \multirow{5}{*}{Transformer}
        & Hidden dimension & 512 \\
        & Attention heads & 8 \\
        & Feed-forward dimension & 3200 \\
        & Encoder / decoder layers & 4 / 7 \\
        & Dropout & 0.1 \\
        \midrule
        \multirow{3}{*}{VAE and actions}
        & Latent dimension & 32 \\
        & KL weight & 10.0 \\
        & Chunk size & 30 to 100 depending on checkpoint \\
        \bottomrule
    \end{tabular}
    \end{adjustbox}
\end{table}

\begin{table}[htbp]
    \centering
    \footnotesize
    \caption{Training configuration for action-chunk policies.}
    \label{tab:app_train}
    \begin{adjustbox}{max width=\linewidth}
    \begin{tabular}{ll}
        \toprule
        Hyperparameter & Value \\
        \midrule
        Optimizer & AdamW \\
        Learning rate & $1\times10^{-5}$ \\
        Backbone learning rate & $1\times10^{-5}$ \\
        Weight decay & $1\times10^{-4}$ \\
        Batch size & 32 \\
        \bottomrule
    \end{tabular}
    \end{adjustbox}
\end{table}

\begin{table}[htbp]
    \centering
    \scriptsize
    \setlength{\tabcolsep}{2.2pt}
    \caption{Controller and haptic guidance parameters corresponding to the main-text notation. Obstacle haptics are rendered on the pedal $x/y$ axes. Yaw remains an input axis but does not receive lidar haptic offsets.}
    \label{tab:app_control}
    \begin{adjustbox}{max width=\linewidth}
    \begin{tabular}{p{0.18\linewidth}p{0.27\linewidth}p{0.22\linewidth}p{0.25\linewidth}}
        \toprule
        Category & Parameter & Value & Main-text role \\
        \midrule
        \multirow{8}{*}{Pedal/obstacle}
        & Pedal position clamp $q^p_x,q^p_y$ & $\pm0.315~\mathrm{rad}$ & Clamp before base-velocity mapping \\
        & Pedal yaw clamp $q^p_\psi$ & $\pm0.5~\mathrm{rad}$ & Clamp yaw input axis \\
        & Deadband $(x,y,\psi)$ & $0.02,0.02,0.075~\mathrm{rad}$ & Zero small foot motion before mapping \\
        & Velocity map limits $(x,y,\psi)$ & $1.0,1.0,1.5708$ & Linear map to m/s, m/s, rad/s \\
        & $r_{\min},r_{\mathrm{far}}$ & $0.40,0.50~\mathrm{m}$ & Active range in $w(r)$ \\
        & $w_{\max},\delta,\alpha$ & $1.0,0.01~\mathrm{m},0.1$ & Weight scale, smoothstep width, filter \\
        & $\eta_{x,y}$ and $p_{x,y}^{\pm}$ & $0.1$ and $\pm0.1~\mathrm{rad}$ & Switch filtered obstacle cue to pedal spring center \\
        & Pedal $K^p_{x,y},K^d_{x,y}$ & $20,2$ & Spring-damper rendering in $\tau^p_{a,t}$ \\
        \midrule
        \multirow{7}{*}{Arm reflection}
        & Follower arm $K_p,K_d$ & $7.0,0.8$ & Position/MIT follower command tracking \\
        & Leader arm $K_p$ & [0.05, 0.1, 0.1, 0.05, 0.1, 0.1] & Low-gain leader MIT control \\
        & Leader arm $K_d$ & [0.1, 0.1, 0.01, 0.1, 0.1, 0.1] & Low-gain leader damping \\
        & Effort filter $\alpha_{\tau}$ & $0.5$ & Filtered follower effort $\boldsymbol{\tau}^{r}_t$ \\
        & Gravity scaling in $\hat{\bb{g}}$ & joints 1 to 3 use $1/4$, joints 4 to 6 use $1$ & Pinocchio gravity term used in reflection \\
        & Torque scale $\bb{s}_{\tau}$ & [0.5, 0.15, 0.6, 0.5, 0.15, 0.6] & Per-joint residual-load scale \\
        & Leader torque limit $\tau_{\max}^{\mathrm{lead}}$ & $18~\mathrm{N\,m}$ & Clip final feedforward torque \\
        \midrule
        \multirow{4}{*}{Reachability}
        & $m_{\min}$ & $0.175$ & Activate when manipulability is low \\
        & $r_{\mathrm{stretch}}$ & $0.6~\mathrm{m}$ & Require stretched arm before assistance \\
        & $v_{\max}$ & $0.05~\mathrm{m/s}$ & Bound additive base-intent velocity \\
        & $w_e,w_g$ & $1.0,0.2$ & Radial and gradient terms in $\bb{d}_i$ \\
        \bottomrule
    \end{tabular}
    \end{adjustbox}
\end{table}

\section{Supplementary Visual Evidence and Enlargened Images}
\label{app:visual_material}

This section keeps only supplementary images that add distinct evidence beyond the main figures. Images are grouped by the hidden variable or evaluation setting they explain, including base clearance under occlusion, reachability, sustained contact, action-chunk policy rollout context, and long-horizon teleoperation.

\subsection{Hardware And Guidance Signals}
\label{app:visual_system}

\begin{figure}[htbp]
    \centering
    \includegraphics[width=0.84\linewidth]{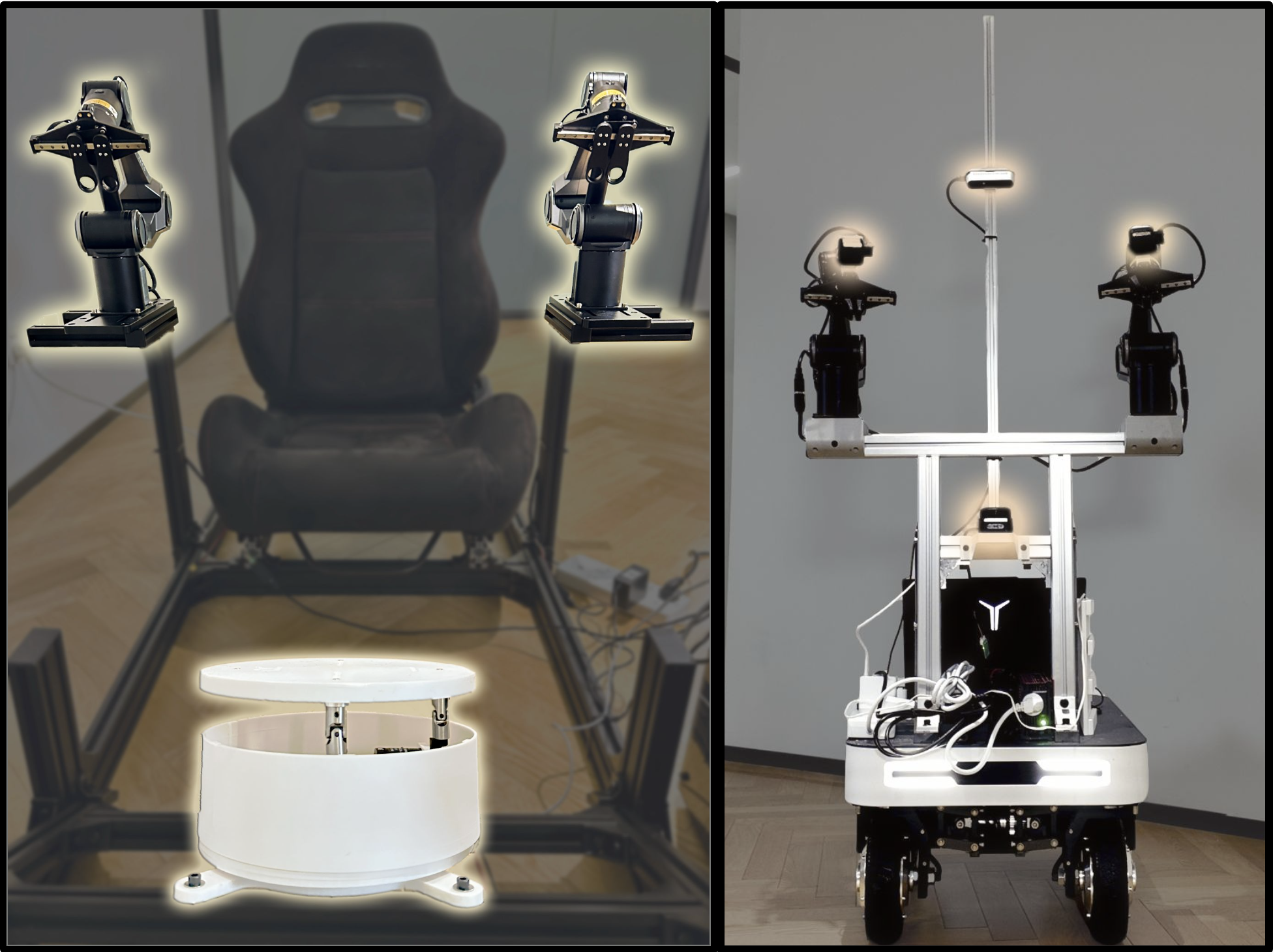}
    \caption{Physical TriPilot-FF hardware used for data collection. The operator station combines two leader arms with the motorized foot pedal, while the robot platform combines an omnidirectional base, two follower arms, wrist cameras, a head camera, and lidar. This complements the main system diagram by showing the actual hardware layout.}
    \label{fig:app_overview_extra}
\end{figure}

\begin{figure}[htbp]
    \centering
    \includegraphics[width=0.33\linewidth]{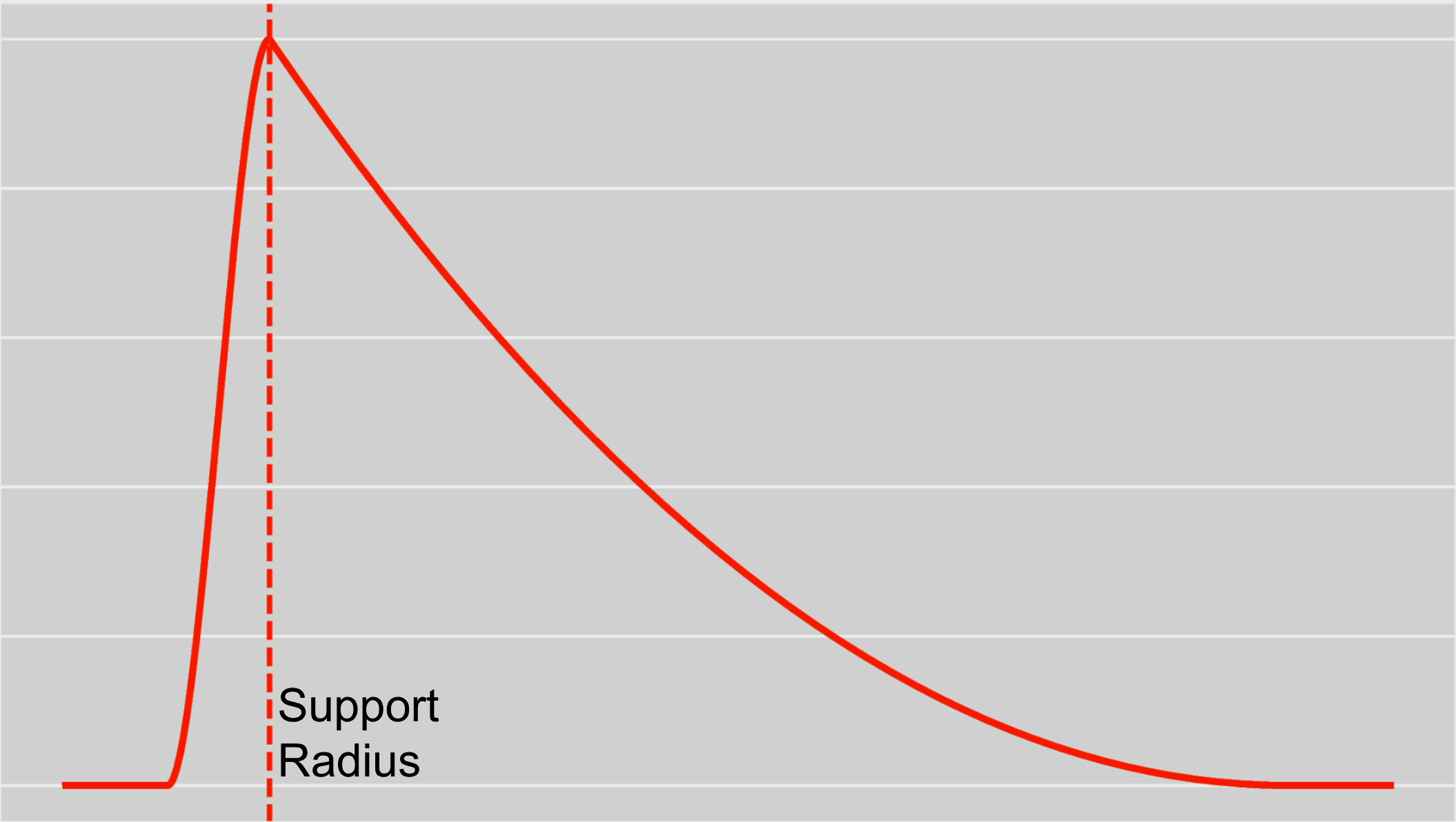}
    \includegraphics[width=0.63\linewidth]{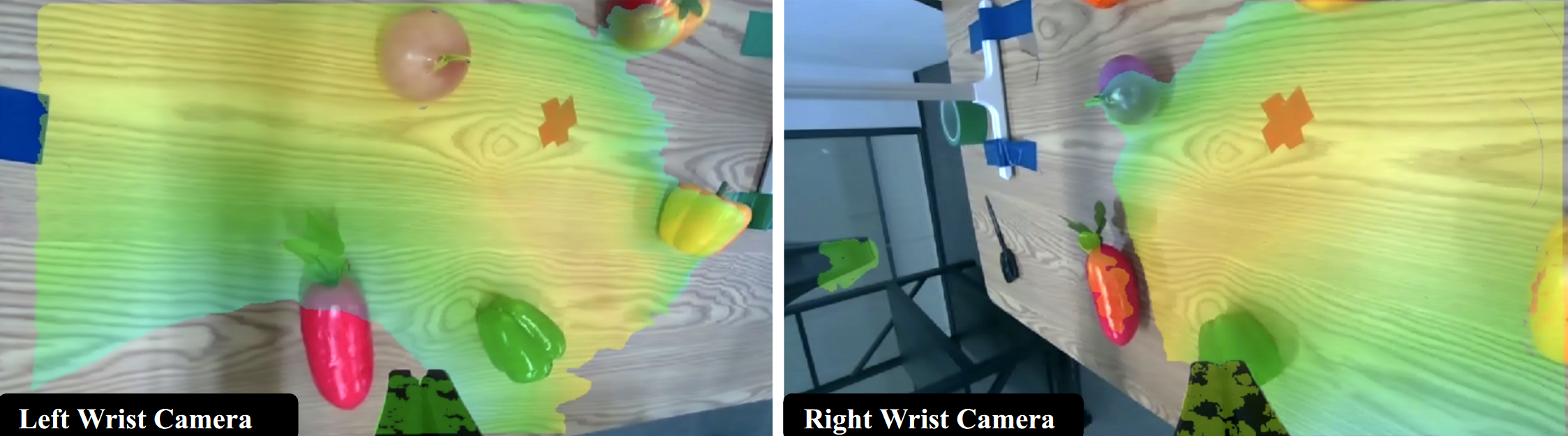}
    \caption{Guidance quantities rendered through the interface. The top plot visualizes the obstacle proximity weighting used by the pedal resistance channel. The bottom row shows wrist-camera visualizations of reachability and manipulability cues, which help the operator recognize when a base reposition would improve arm conditioning.}
    \label{fig:app_guidance_quantities}
\end{figure}

\begin{figure}[htbp]
    \centering
    \includegraphics[width=0.88\linewidth]{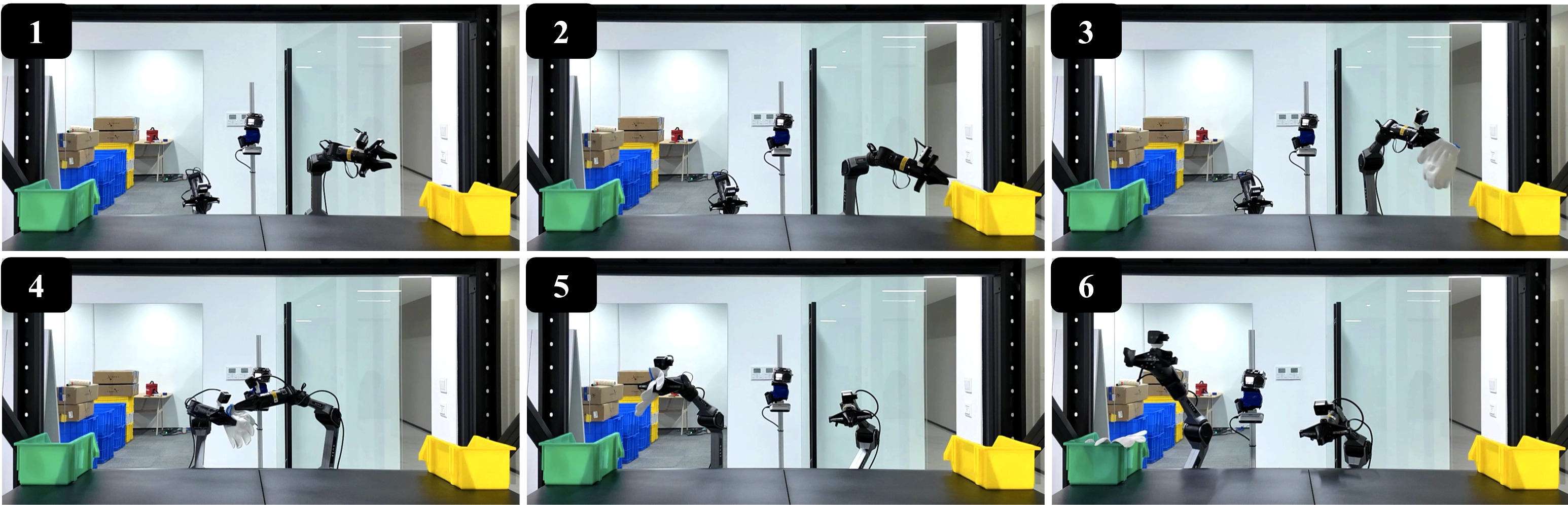}
    \caption{GuidedReach sequence under manipulability guidance. As the active arm approaches an awkward stretched posture, the reachability node encourages a small base reposition before the arm remains in a low-dexterity configuration. The sequence illustrates the intended log effect of earlier base motion and less low-manipulability dwell.}
    \label{fig:app_guided_reach_sequence}
\end{figure}

The first visual block connects the physical interface to the cues used in the experiments. The hardware photograph in \cref{fig:app_overview_extra} fixes the spatial relationship between the operator, pedal, leader arms, and robot, while \cref{fig:app_guidance_quantities,fig:app_guided_reach_sequence} show the signals that turn this setup into a demonstration-shaping interface.

These images should be read as qualitative evidence for the feedback channels rather than as additional metrics. The proximity plot explains why the pedal becomes resistive only in the configured near-obstacle band, and the wrist-camera overlays illustrate how a low-manipulability posture is made visible before the operator commits to a poor arm configuration.

\subsection{Occlusion And Base Clearance}
\label{app:visual_navigation}

\begin{figure}[htbp]
    \centering
    \includegraphics[width=0.84\linewidth]{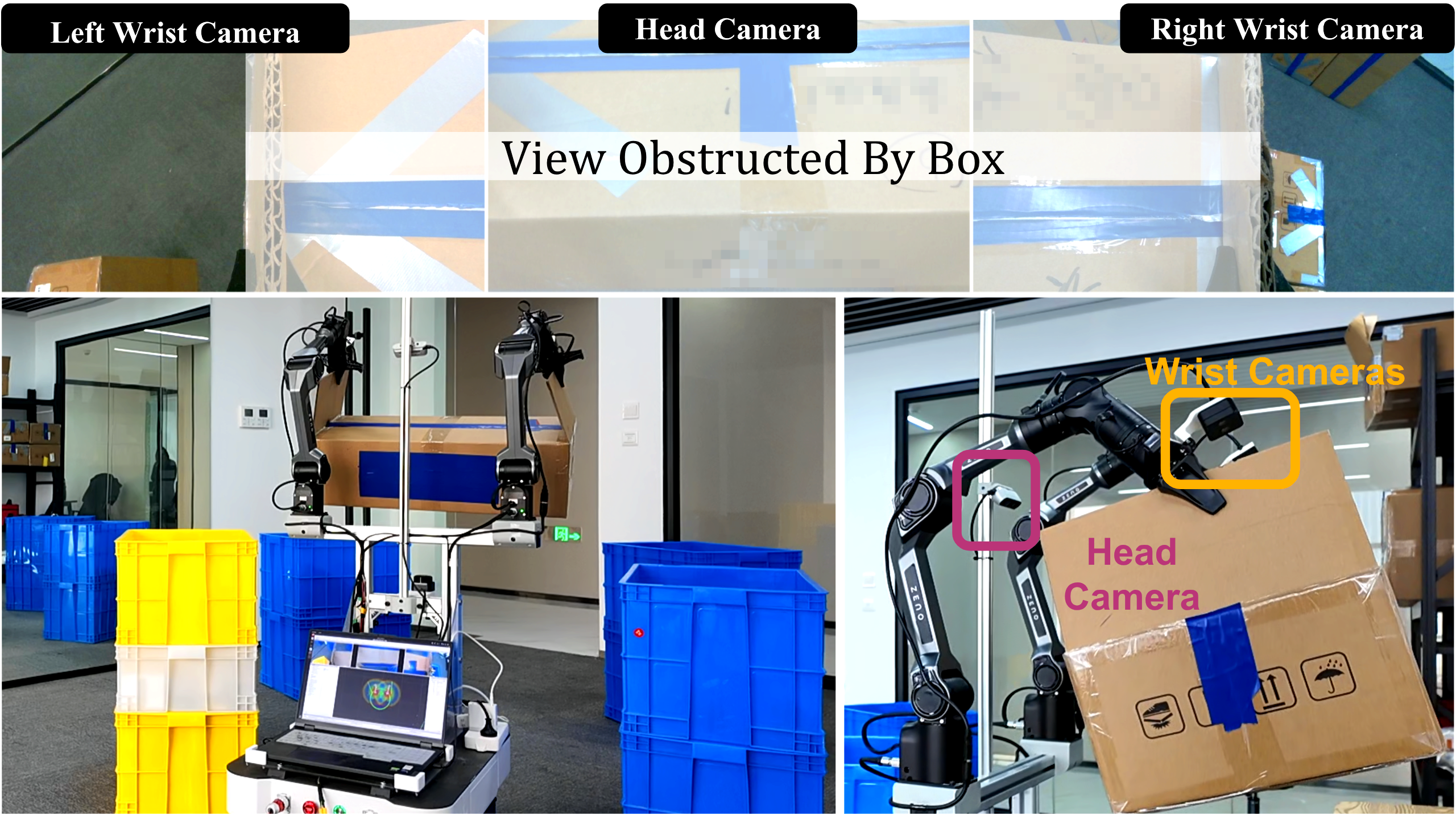}
    \caption{BlindCarry sensing problem. The carried box blocks the head-camera view of nearby obstacles, while the wrist cameras mainly see the grasp and object surface. The lidar-derived clearance signal is therefore rendered at the pedal so that collision risk is felt where the base command is selected.}
    \label{fig:app_obstruction_extra}
\end{figure}

\begin{figure}[htbp]
    \centering
    \includegraphics[width=0.84\linewidth]{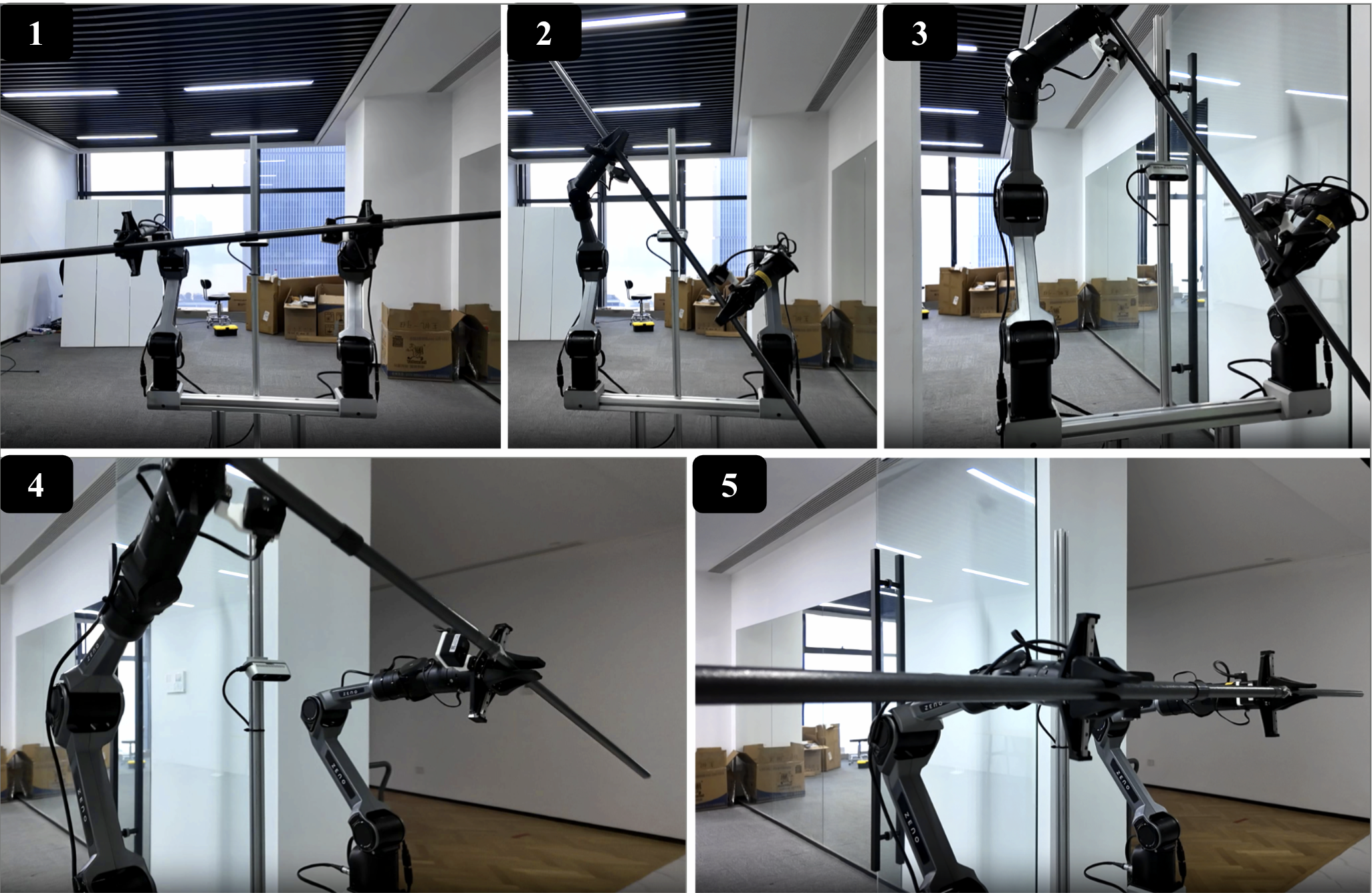}
    \caption{NarrowTransport setup and keyframes. The long object creates a swept volume that is difficult to judge from a single camera view, so the task stresses coordinated base motion, arm posture, and obstacle-aware pedal resistance.}
    \label{fig:app_narrow_passage_extra}
\end{figure}

The navigation examples separate two cases that are easy to conflate in a montage. \Cref{fig:app_obstruction_extra} explains why visual teleoperation can miss nearby obstacles under object occlusion. \Cref{fig:app_narrow_passage_extra} then shows a different failure mode, where the object itself increases the swept volume and makes small base-angle errors costly.

\begin{figure}[htbp]
    \centering
    \includegraphics[width=0.48\linewidth]{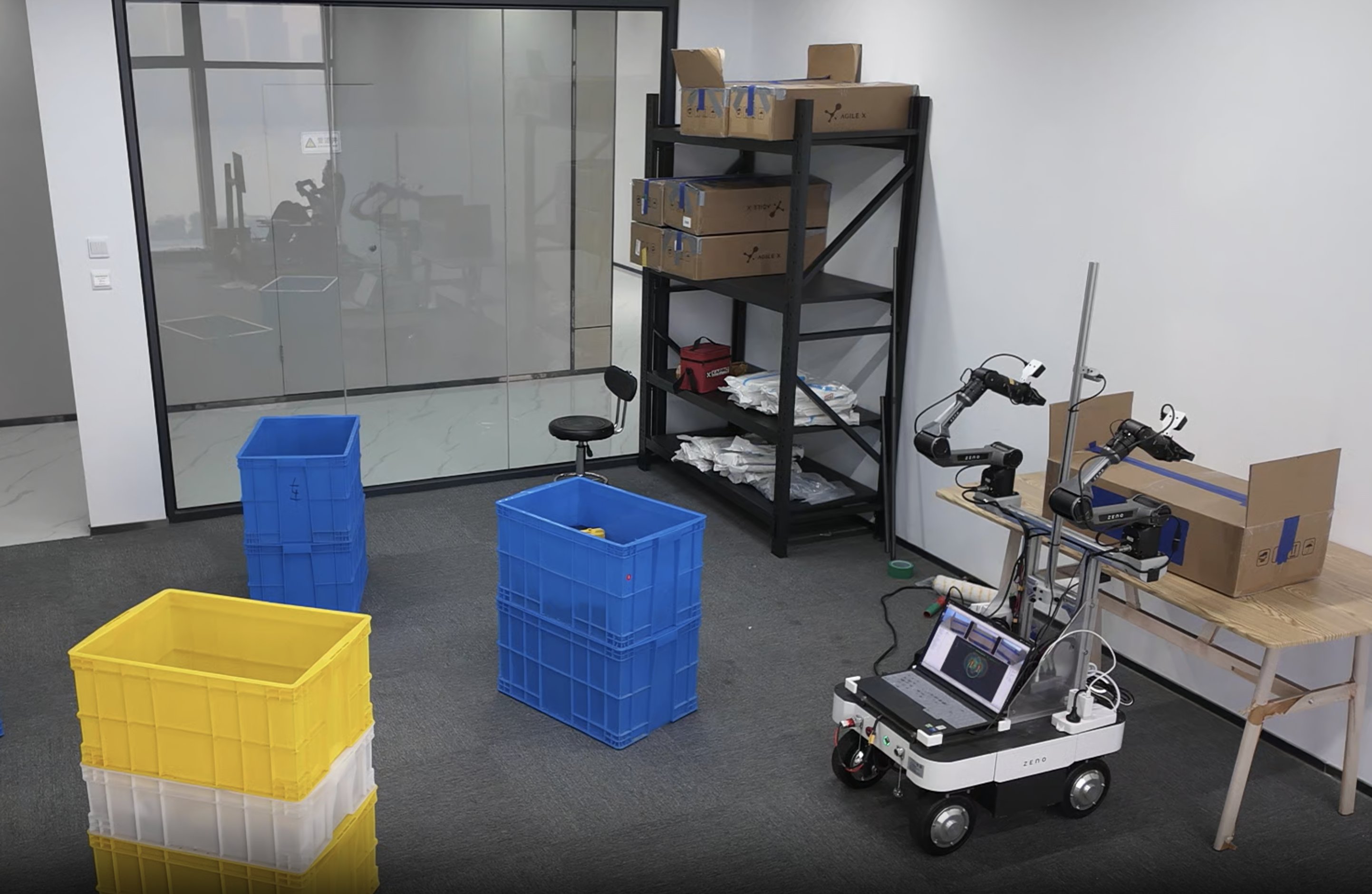}
    \hfill
    \includegraphics[width=0.48\linewidth]{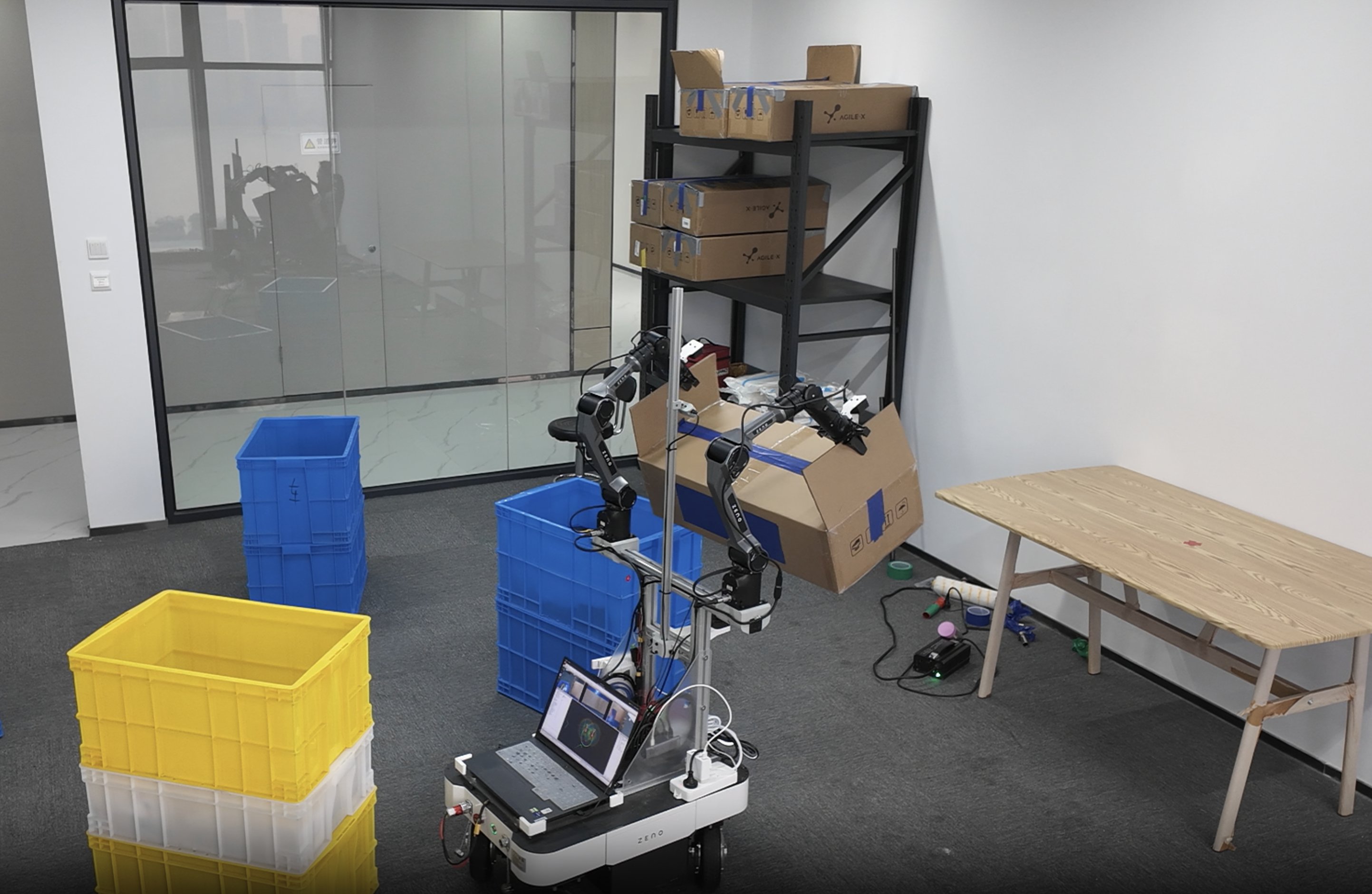}\\[1pt]
    \includegraphics[width=0.48\linewidth]{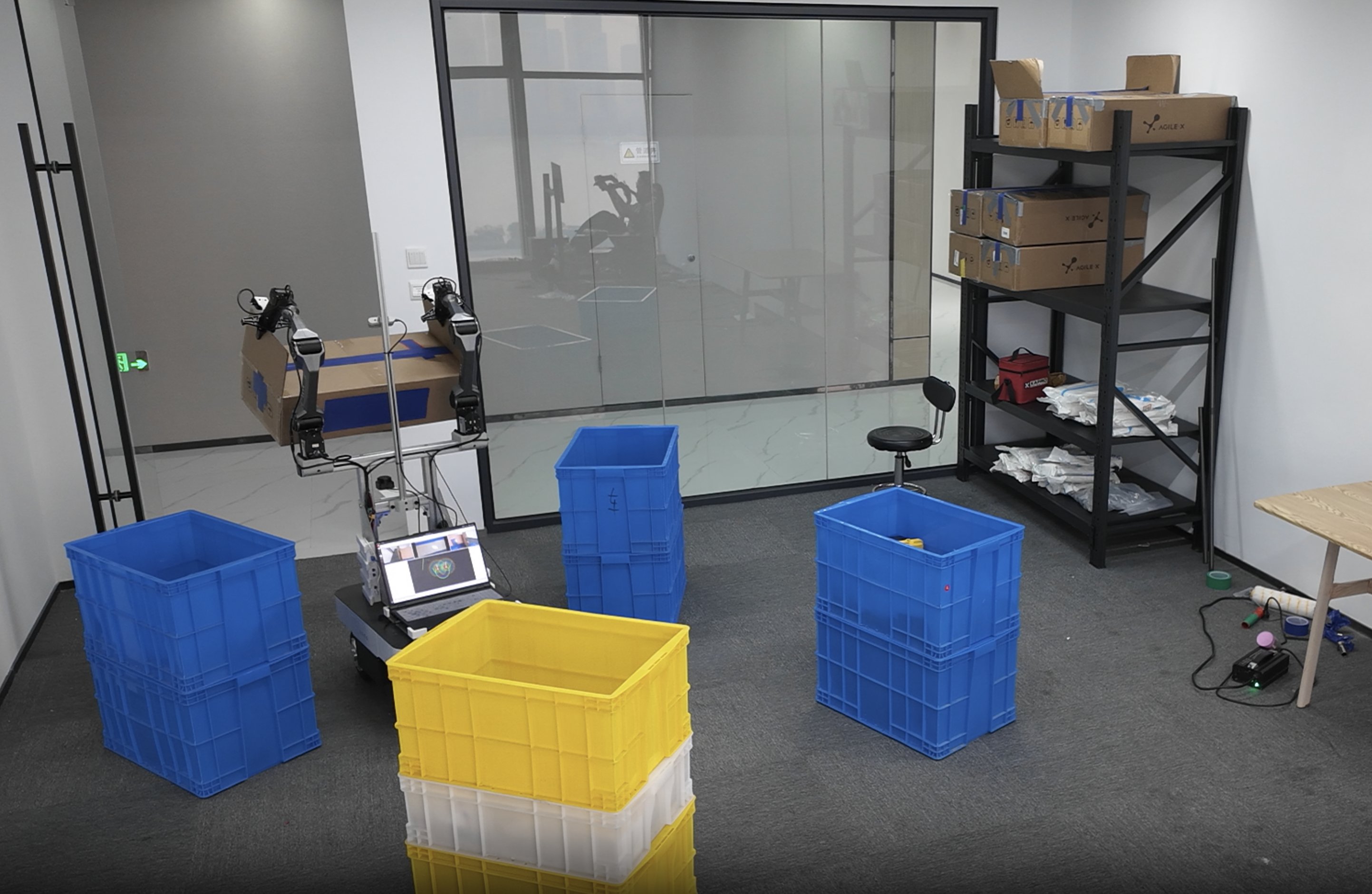}
    \hfill
    \includegraphics[width=0.48\linewidth]{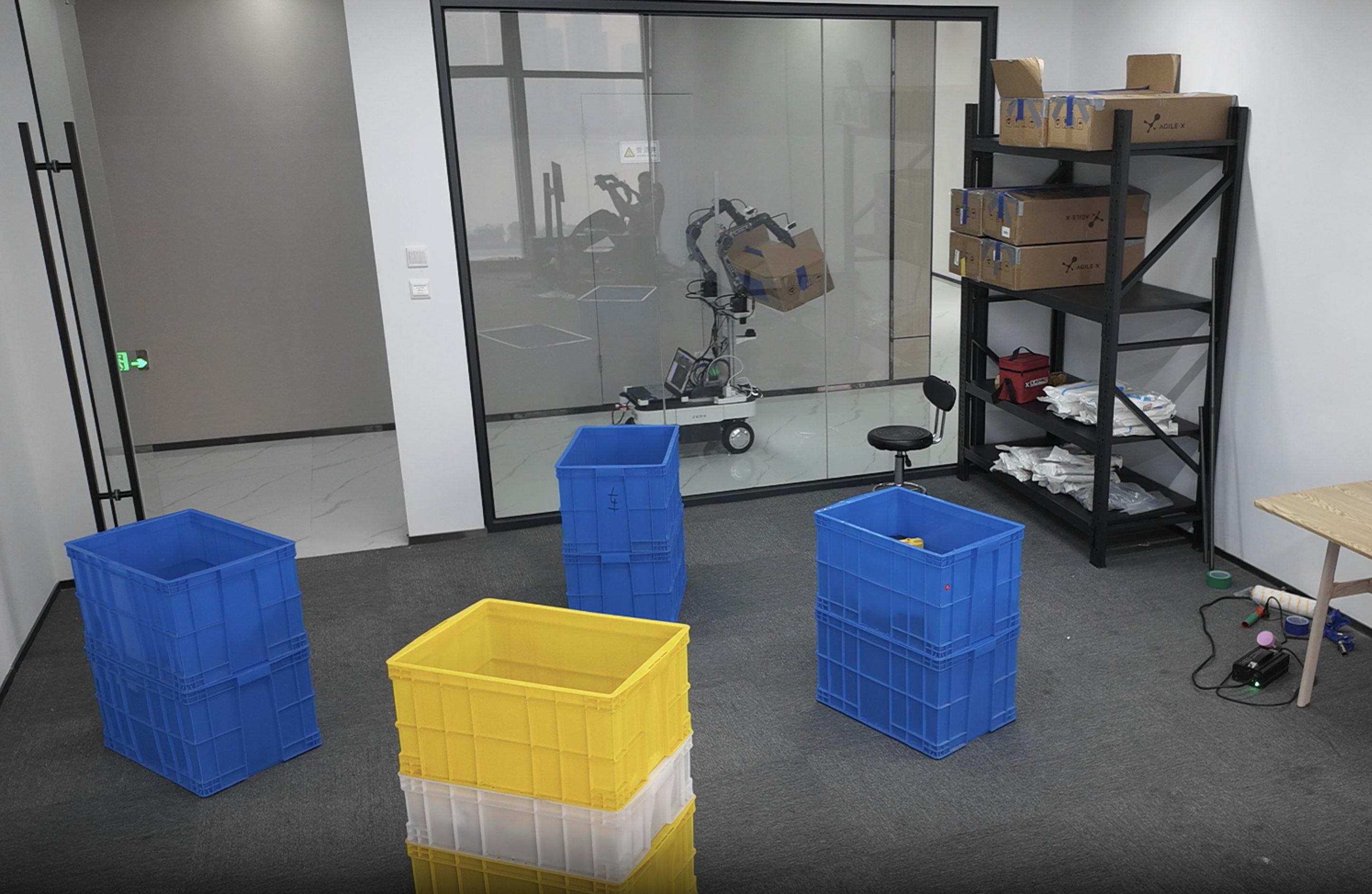}
    \caption{BlindCarry keyframes. The robot transports a bulky box through scattered bins, with the object obscuring the most relevant clearance region. The selected frames show the approach, entry into clutter, close-clearance turn, and exit, avoiding redundant intermediate frames from the same run.}
    \label{fig:app_avoid_sequence}
\end{figure}

The four BlindCarry frames are intentionally sparse. They preserve the stages needed to interpret the collision and timing metrics, namely approach, entry, close-clearance turn, and exit, without turning the appendix into a frame-by-frame video dump.

\subsection{Contact-Rich Manipulation}
\label{app:visual_tasks}

\begin{figure}[htbp]
    \centering
    \includegraphics[width=0.84\linewidth]{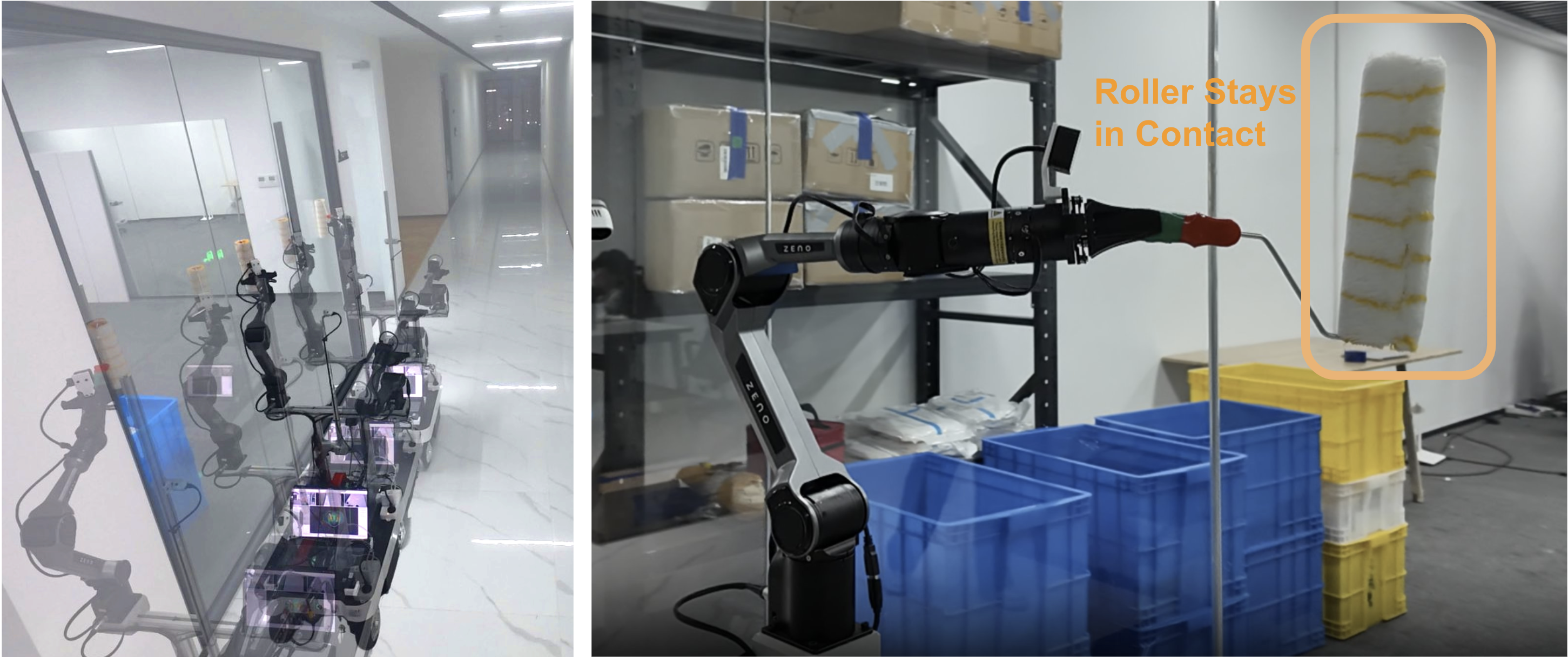}
    \caption{MobileSwipe contact setup. The robot maintains roller contact while moving the base laterally. This setting motivates leader-arm force reflection because contact loss or excessive normal force can occur before it is obvious in RGB frames.}
    \label{fig:app_roller_contact_extra}
\end{figure}

\begin{figure}[htbp]
    \centering
    \includegraphics[width=0.48\linewidth]{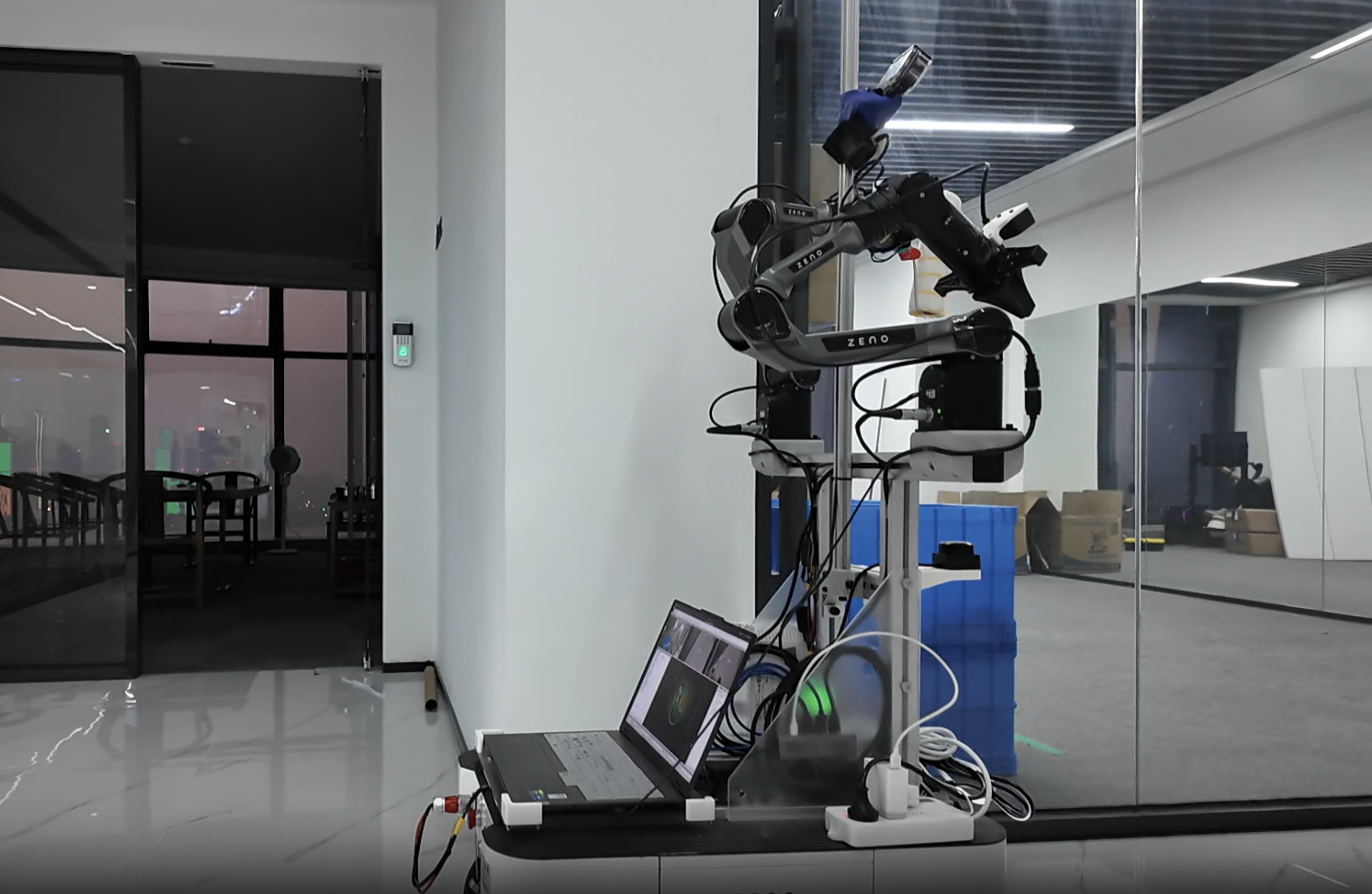}
    \hfill
    \includegraphics[width=0.48\linewidth]{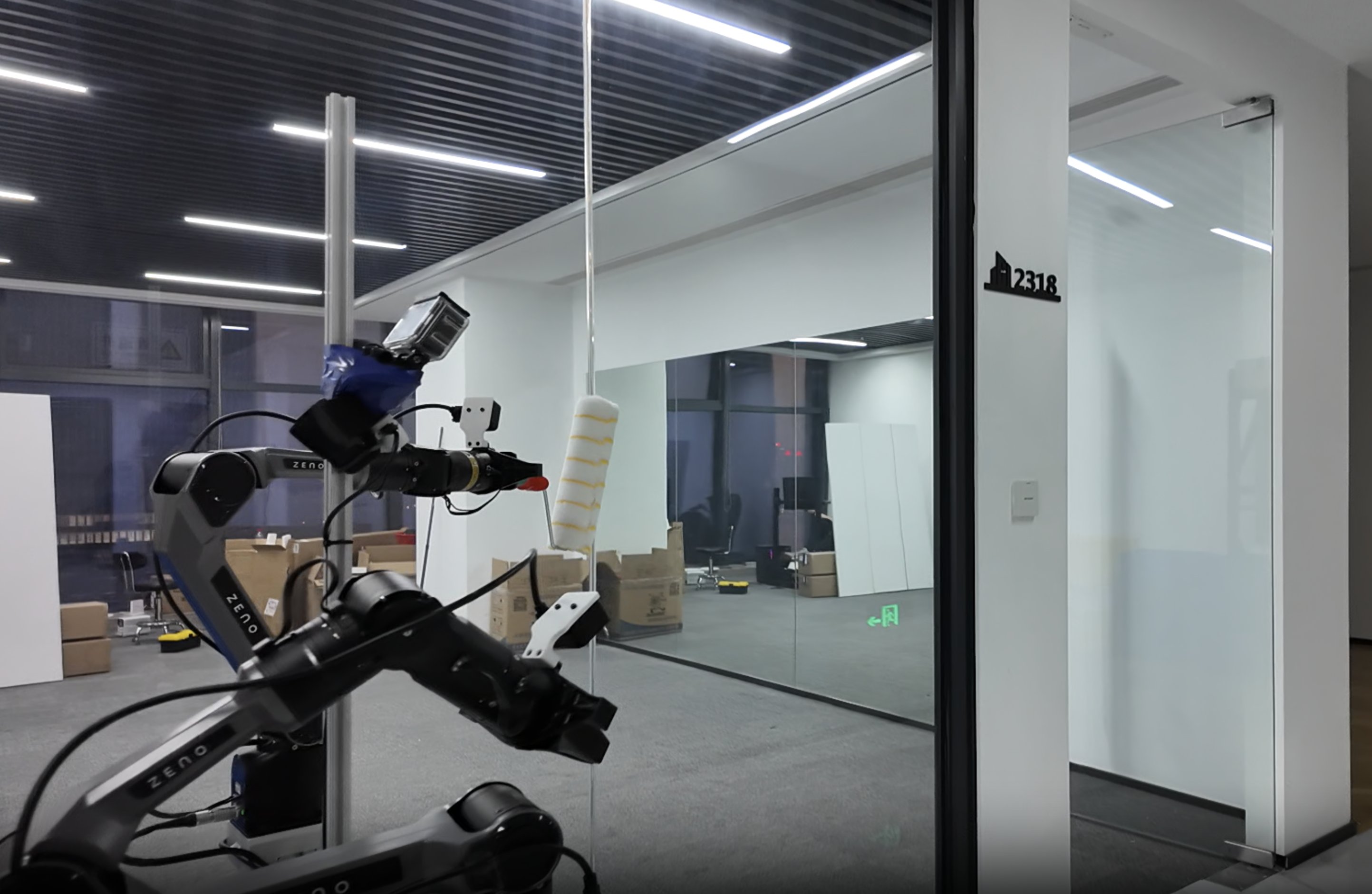}\\[1pt]
    \includegraphics[width=0.48\linewidth]{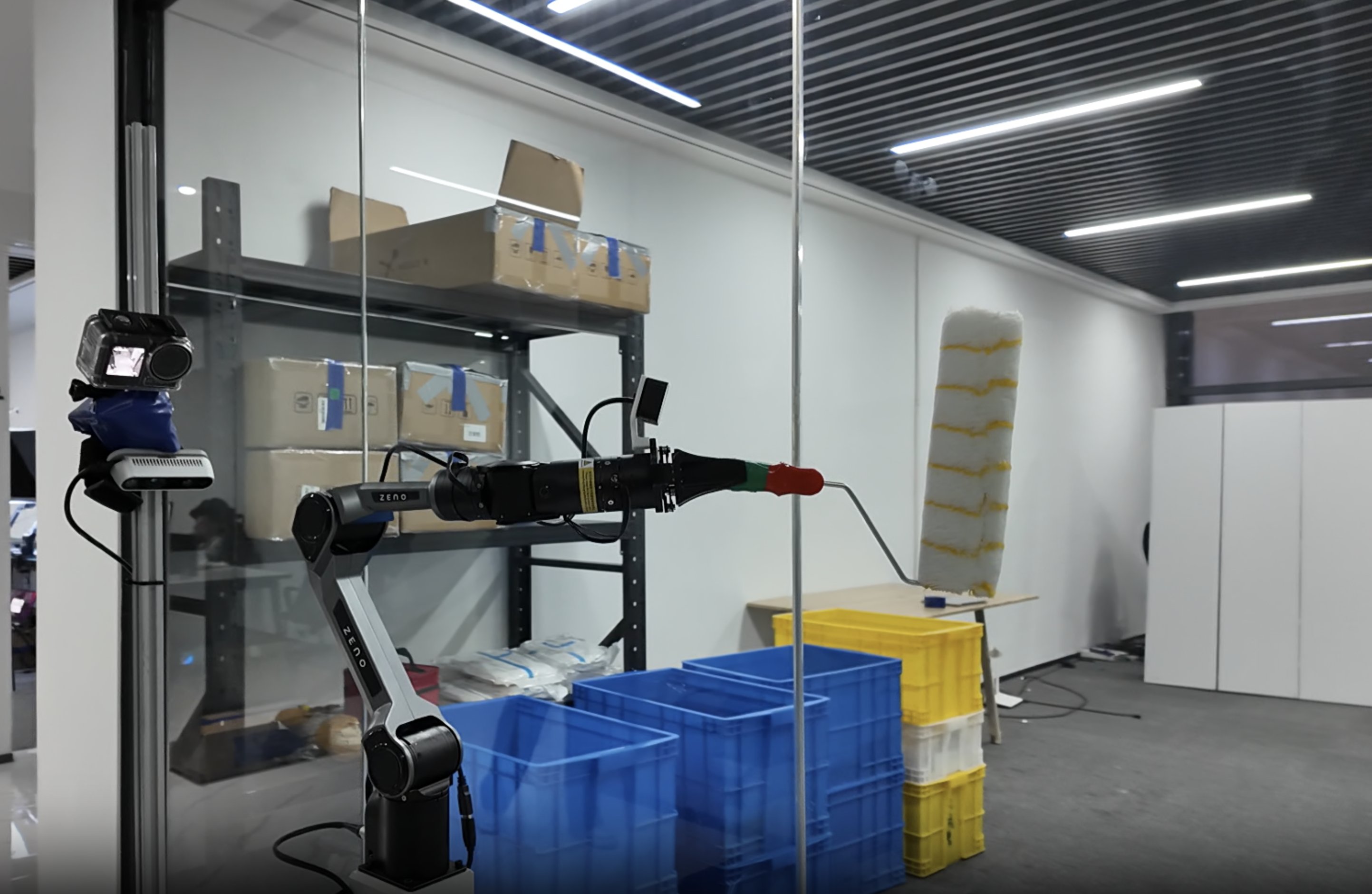}
    \hfill
    \includegraphics[width=0.48\linewidth]{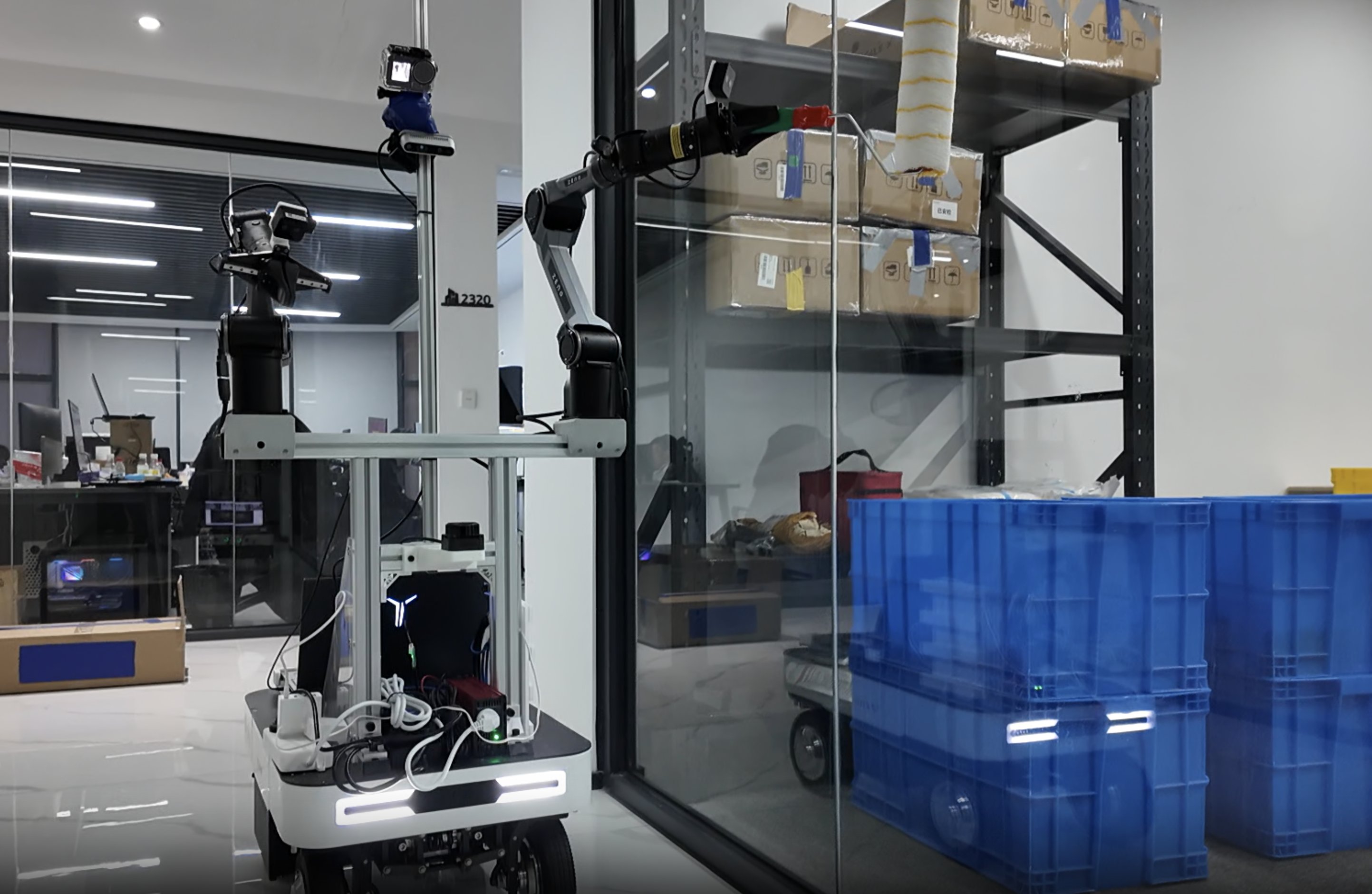}
    \caption{MobileSwipe keyframes. The sequence shows approach, contact establishment, lateral sliding under contact, and completion. Only four frames are retained because the omitted frames repeat the same continuous contact phase.}
    \label{fig:app_swipe_sequence}
\end{figure}

The contact-rich examples focus on events whose timing is not reliably available from RGB alone. The setup view in \cref{fig:app_roller_contact_extra} identifies the maintained contact patch, and the keyframes in \cref{fig:app_swipe_sequence} show the phases where torque residuals help distinguish smooth sliding from contact loss or excessive normal loading.

\subsection{Policy Rollout Task Views}
\label{app:visual_policy_tasks}

\begin{figure}[htbp]
    \centering
    \includegraphics[width=0.88\linewidth]{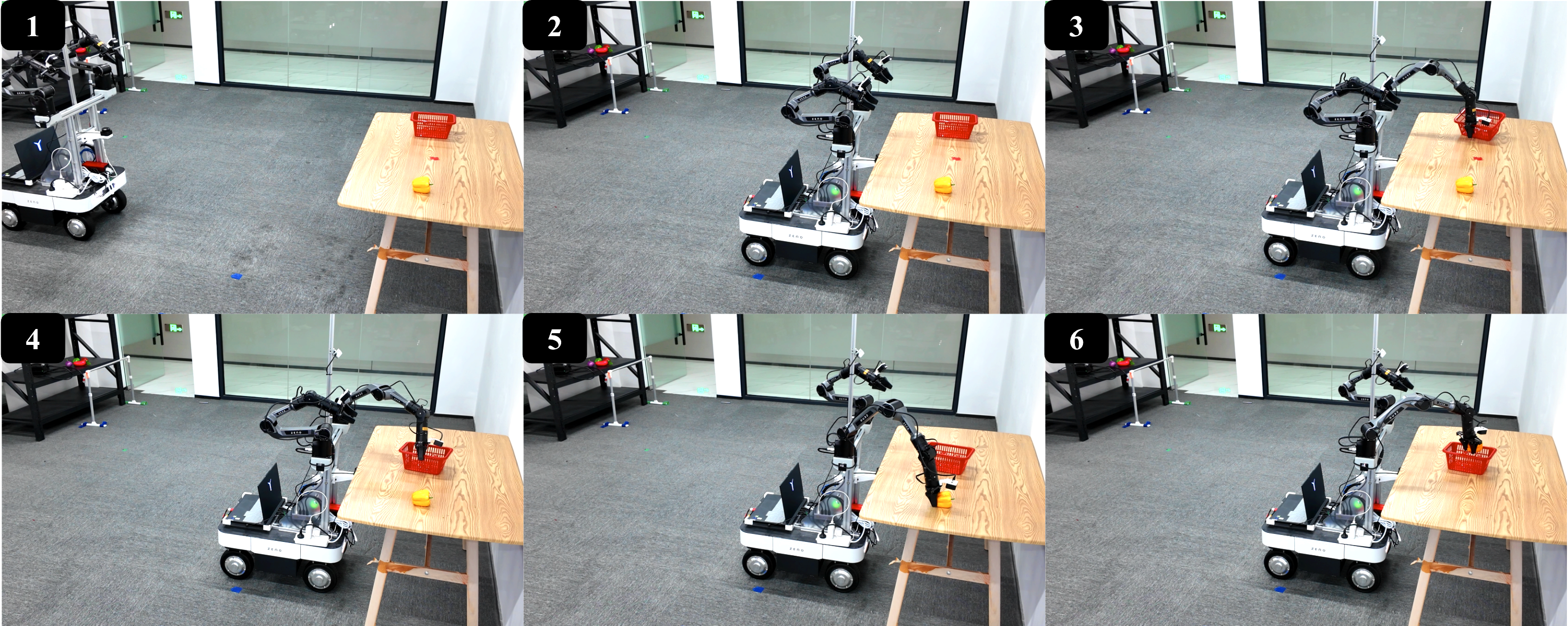}
    \caption{BasketPack rollout view. The sequence combines base approach, basket interaction, and object placement. It is one of the real-world fixed-dataset action-chunk policy evaluations for the deployed visual-proprioceptive whole-body action expert.}
    \label{fig:app_basket_extra}
\end{figure}

The policy-rollout panels provide task context for the fixed-dataset tables rather than new quantitative claims. They are grouped here so the reader can compare the kind of hidden constraint each rollout stresses, including basket interaction, hanger extraction and transfer, deformable laundry handling, and the simplified simulation contacts.

\begin{figure}[htbp]
    \centering
    \includegraphics[width=0.88\linewidth]{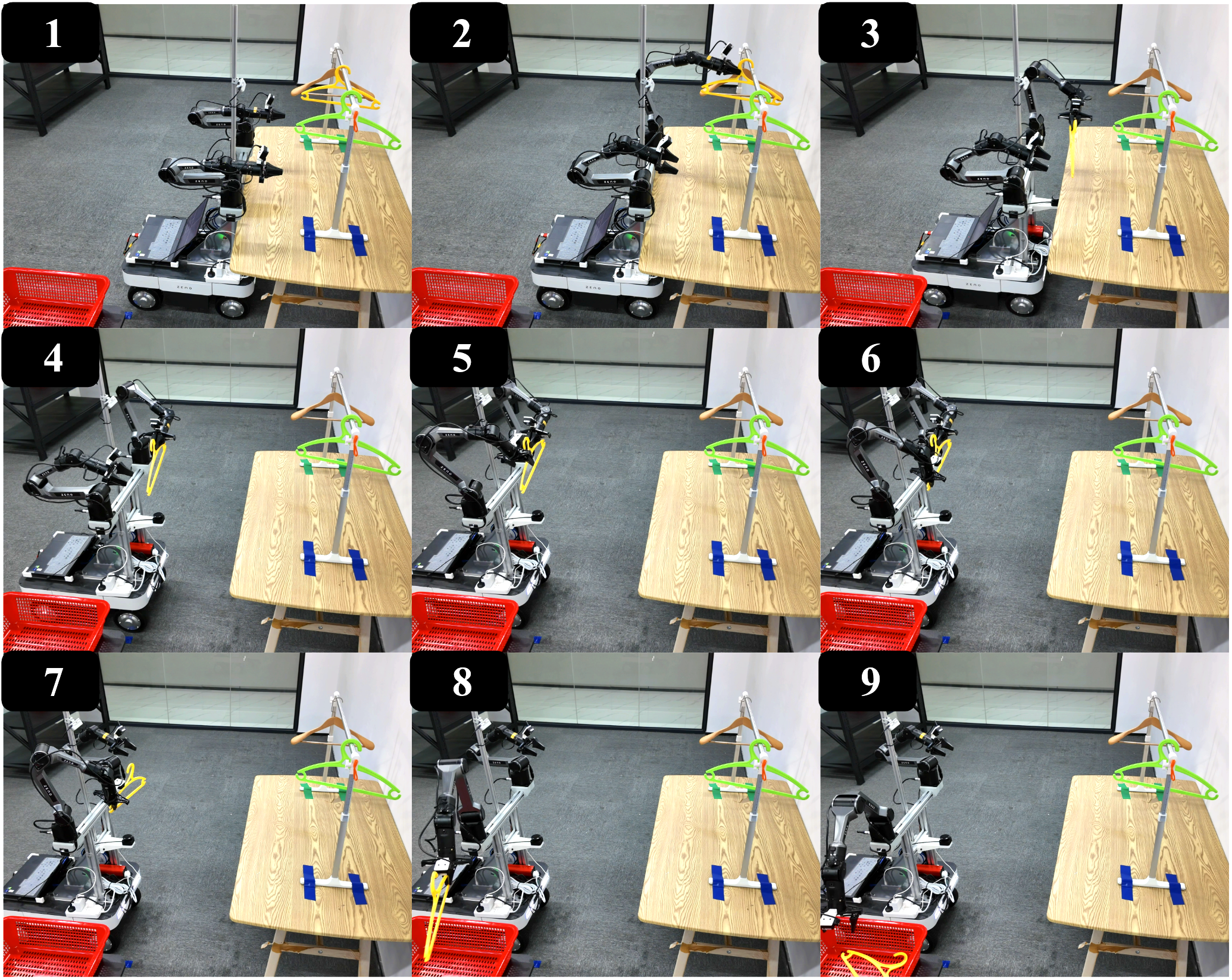}
    \caption{HangerHandOff rollout sequence. The robot extracts a hanger from a rack, performs a bimanual transfer, and moves the object toward the target container. This single sequence is kept instead of multiple visually similar hanger montages because it captures the key contact events, including rack extraction, bimanual load sharing, and release timing.}
    \label{fig:app_hanger_sequence}
\end{figure}

For HangerHandOff, the important visual transitions are not just object pose changes. The rack extraction and bimanual transfer phases mark moments where contact state is ambiguous in RGB. In the present paper, effort history is used to analyze these events, while the deployed policy itself uses visual-proprioceptive state and predicts whole-body action chunks.

\begin{figure}[htbp]
    \centering
    \includegraphics[width=0.90\linewidth]{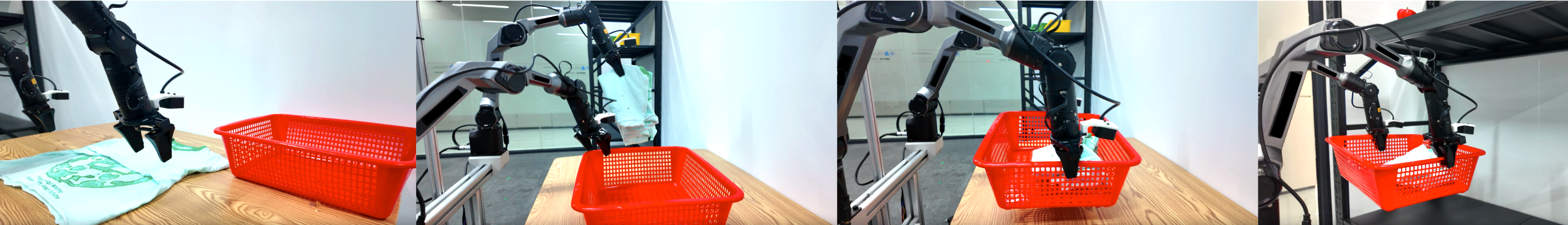}
    \caption{LaundryTransport rollout view. The sequence shows cloth folding, basket packing, and basket placement stages, providing the representative laundry task view for the policy rollout appendix.}
    \label{fig:app_laundry_policy_extra}
\end{figure}

\begin{figure}[htbp]
    \centering
    \includegraphics[width=0.90\linewidth]{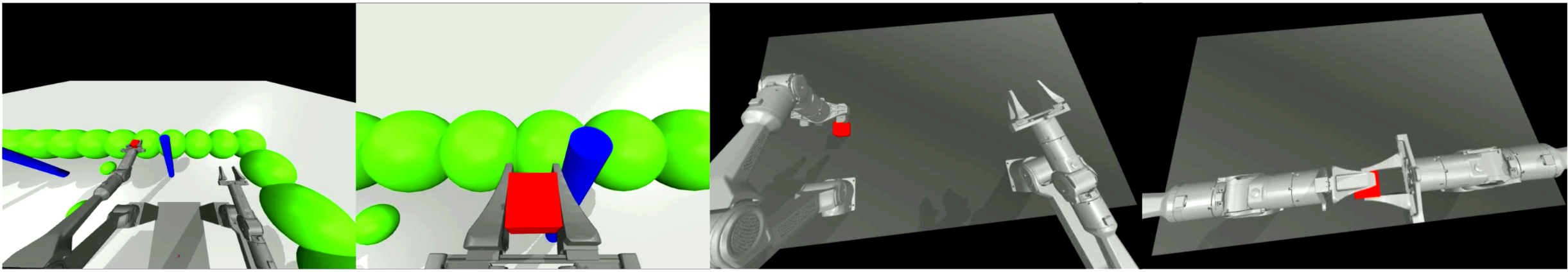}
    \caption{Simulation policy-evaluation view. The simulated transfer and placement tasks are used for fixed-dataset action-chunk policy evaluations with visual and proprioceptive observations.}
    \label{fig:app_sim_policy_extra}
\end{figure}

\subsection{Long-Horizon Scene Views}
\label{app:visual_long_horizon}

The long-horizon views are kept as scene summaries. They show that the same operator interface is used continuously across navigation, reaching, object handling, and contact-rich subtasks, which is why the corresponding table reports full-scene completion and intervention counts instead of isolated frame-level outcomes.

\begin{figure}[htbp]
    \centering
    \includegraphics[width=0.90\linewidth]{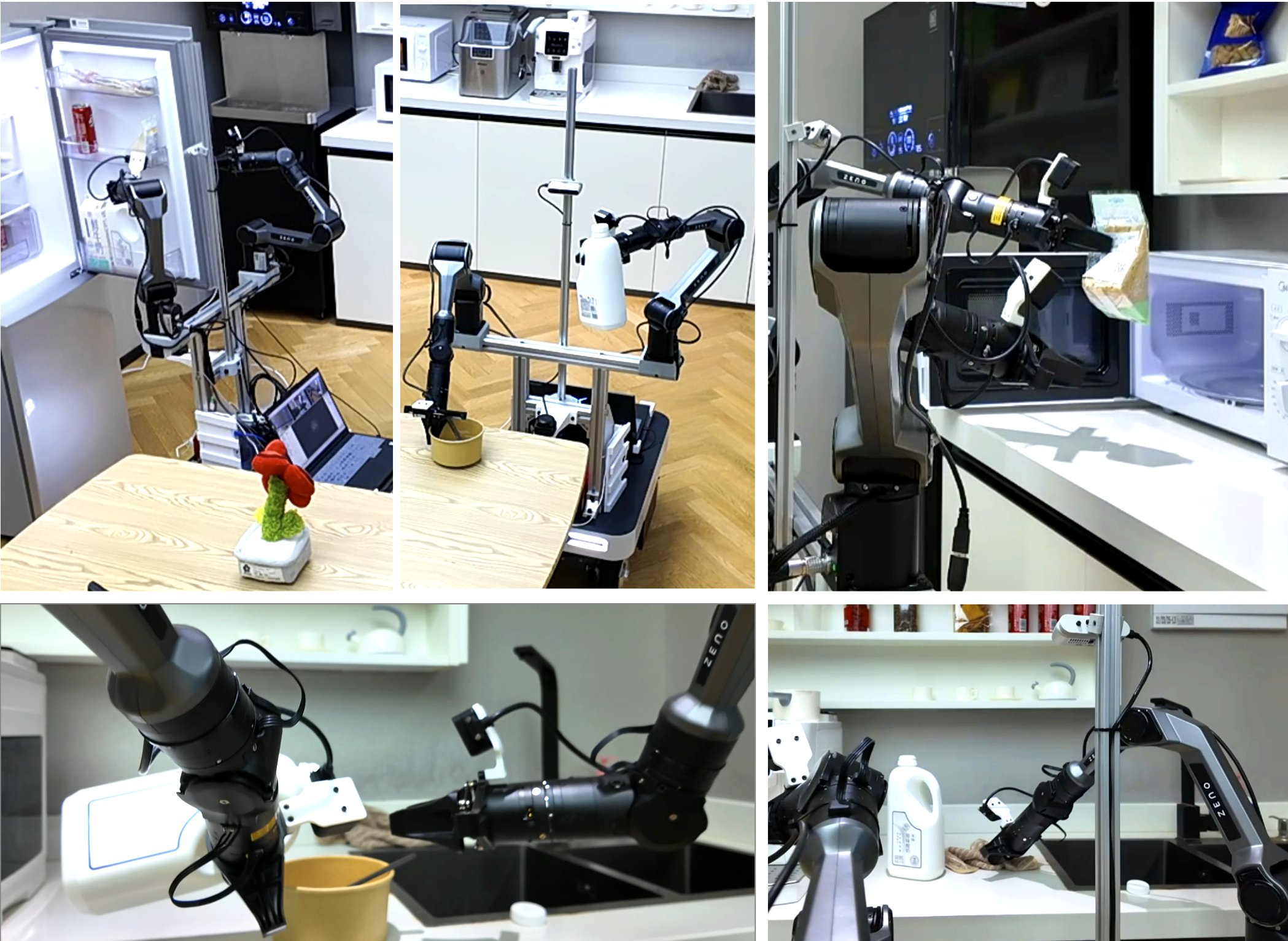}
    \caption{Kitchen long-horizon teleoperation scene. The montage covers navigation and manipulation around appliances, containers, pouring, and wiping. The episode is executed continuously without resetting robot pose or interface state.}
    \label{fig:app_kitchen_extra}
\end{figure}

\begin{figure}[htbp]
    \centering
    \includegraphics[width=0.90\linewidth]{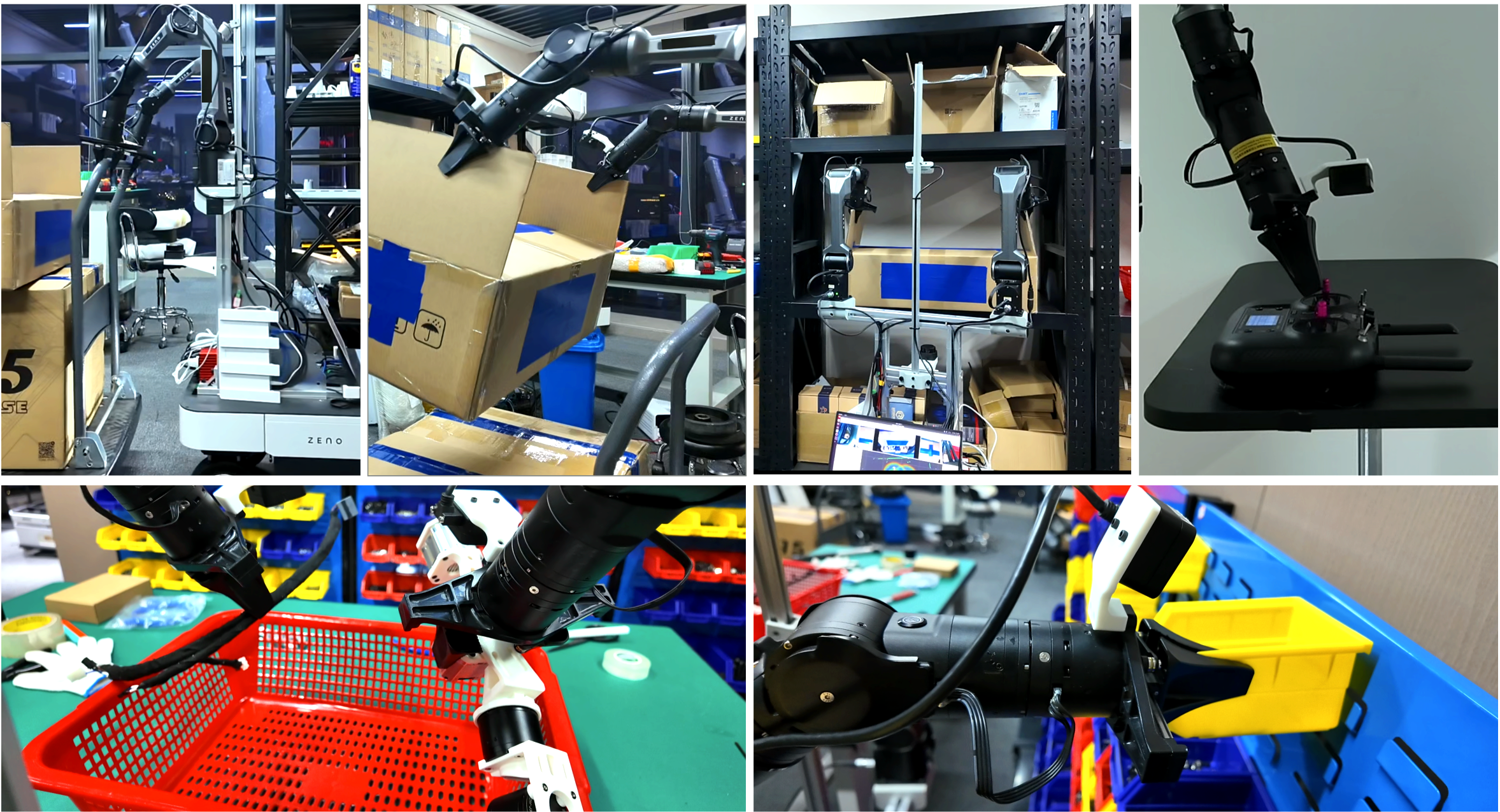}
    \caption{Tool-room long-horizon teleoperation scene. The montage includes shelf access, close-clearance object handling, packing, and tool and device interaction, matching the repeated room-scale readout in the main paper.}
    \label{fig:app_toolroom_extra}
\end{figure}

% \clearpage
% \section{Supplementary Video Guidance}
% \label{app:video}
% The supplementary video should make the teleoperation mechanism visible rather than only showing robot-side outcomes. Recommended panels are operator foot motion, leader-arm motion, synchronized haptic and contact signals, and the robot view. For ablations, the reduced-feedback and full-feedback trials should be shown side by side with identical scales. For policy rollouts, the video should show matched initial states for the deployed policy, with first blocking failures labeled when a rollout fails. The long-horizon segments should display the completed statistics reported in the main paper. Kitchen uses 5 trials, 80.0\% full success, and 0.2 interventions per run. Tool room uses 5 trials, 100.0\% full success, and 1.2 close-clearance events per run.

%% file: ref.bib
@incollection{HART1988139,
title = {Development of NASA-TLX (Task Load Index): Results of Empirical and Theoretical Research},
editor = {Peter A. Hancock and Najmedin Meshkati},
series = {Advances in Psychology},
volume = {52},
pages = {139-183},
year = {1988},
booktitle = {Human Mental Workload},
author = {Sandra G. Hart and Lowell E. Staveland},
}

@inproceedings{chi2023diffusionpolicy,
	title={Diffusion Policy: Visuomotor Policy Learning via Action Diffusion},
	author={Chi, Cheng and Feng, Siyuan and Du, Yilun and Xu, Zhenjia and Cousineau, Eric and Burchfiel, Benjamin and Song, Shuran},
	booktitle={Proceedings of Robotics: Science and Systems (RSS)},
	year={2023}
}

@inproceedings{Mandlekar2018ROBOTURKAC,
  title={ROBOTURK: A Crowdsourcing Platform for Robotic Skill Learning through Imitation},
  author={Ajay Mandlekar and Yuke Zhu and Animesh Garg and Jonathan Booher and Max Spero and Albert Tung and Julian Gao and John Emmons and Anchit Gupta and Emre Orbay and Silvio Savarese and Li Fei-Fei},
  booktitle={Conference on Robot Learning},
  year={2018}
}

@article{Zhang2017DeepIL,
  title={Deep Imitation Learning for Complex Manipulation Tasks from Virtual Reality Teleoperation},
  author={Tianhao Zhang and Zoe McCarthy and Owen Jow and Dennis Lee and Ken Goldberg and P. Abbeel},
  journal={IEEE International Conference on Robotics and Automation (ICRA)},
  year={2018},
}

@inproceedings{chi2024universal,
	title={Universal Manipulation Interface: In-The-Wild Robot Teaching Without In-The-Wild Robots},
	author={Chi, Cheng and Xu, Zhenjia and Pan, Chuer and Cousineau, Eric and Burchfiel, Benjamin and Feng, Siyuan and Tedrake, Russ and Song, Shuran},
	booktitle={Proceedings of Robotics: Science and Systems (RSS)},
	year={2024}
}

@inproceedings{todorov2012mujoco,
  title={MuJoCo: A physics engine for model-based control},
  author={Todorov, Emanuel and Erez, Tom and Tassa, Yuval},
  booktitle={IEEE/RSJ International Conference on Intelligent Robots and Systems},
  year={2012},
}

@inproceedings{fu2024mobile,
  author    = {Fu, Zipeng and Zhao, Tony Z. and Finn, Chelsea},
  title     = {Mobile ALOHA: Learning Bimanual Mobile Manipulation with Low-Cost Whole-Body Teleoperation},
  booktitle = {{Conference on Robot Learning (CoRL)}},
  year      = {2024},
}

@article{dass2024telemoma,
  title={TeleMoMa: A Modular and Versatile Teleoperation System for Mobile Manipulation},
  author={Dass, Shivin and Ai, Wensi and Jiang, Yuqian and Singh, Samik and Hu, Jiaheng and Zhang, Ruohan and Stone, Peter and Abbatematteo, Ben and Martin-Martin, Roberto},
  journal={arXiv preprint arXiv:2403.07869},
  year={2024}
}

@article{iyer2024openteach,
  title={OPEN TEACH: A Versatile Teleoperation System for Robotic Manipulation},
  author={Iyer, Aadhithya and Peng, Zhuoran and Dai, Yinlong and Guzey, Irmak and Haldar, Siddhant and Chintala, Soumith and Pinto, Lerrel},
  journal={arXiv preprint arXiv:2403.07870},
  year={2024}
}

@article{wu2025robocopilot,
  title={RoboCopilot: Human-in-the-loop Interactive Imitation Learning for Robot Manipulation},
  author={Wu, Philipp and Shentu, Yide and Liao, Qiayuan and Jin, Ding and Guo, Menglong and Sreenath, Koushil and Lin, Xingyu and Abbeel, Pieter},
  journal={arXiv preprint arXiv:2503.07771},
  year={2025}
}

@article{hokayem2006bilateral,
  title={Bilateral teleoperation: An historical survey},
  author={Hokayem, Peter F. and Spong, Mark W.},
  journal={Automatica},
  volume={42},
  number={12},
  pages={2035--2057},
  year={2006}
}

@ARTICLE{Mobile_MPC,
  author={Minniti, Maria Vittoria and Farshidian, Farbod and Grandia, Ruben and Hutter, Marco},
  journal={IEEE Robotics and Automation Letters}, 
  title={Whole-Body MPC for a Dynamically Stable Mobile Manipulator}, 
  year={2019},}

@article{Planning_mobile,
author = {Jindong Tan and Ning Xi and Yuechao Wang},
title ={Integrated Task Planning and Control for Mobile Manipulators},
journal = {The International Journal of Robotics Research},
volume = {22},
number = {5},
year = {2003},
}

@article{TAMP_mobile,
author = {Caelan Reed Garrett and Tomás Lozano-Pérez and Leslie Pack Kaelbling},
title ={FFRob: Leveraging symbolic planning for efficient task and motion planning},

journal = {The International Journal of Robotics Research},
volume = {37},
number = {1},
pages = {104-136},
year = {2018}
}

@INPROCEEDINGS{RT_X,
  author={Abby O’Neill and et al.},
  booktitle={IEEE International Conference on Robotics and Automation (ICRA)}, 
  title={Open X-Embodiment: Robotic Learning Datasets and RT-X Models : Open X-Embodiment Collaboration}, 
  year={2024}
}

@Article{review_mobile,
AUTHOR = {Sandakalum, Thushara and Ang, Marcelo H.},
TITLE = {Motion Planning for Mobile Manipulators—A Systematic Review},
JOURNAL = {Machines},
VOLUME = {10},
YEAR = {2022},
NUMBER = {2}
}

@INPROCEEDINGS{Zhao2023LearningFB,
  title={Learning Fine-Grained Bimanual Manipulation with Low-Cost Hardware},
  author={Tony Zhao and Vikash Kumar and Sergey Levine and Chelsea Finn},
  booktitle={Robotics: Science and Systems (RSS)},
  year={2023},
}

@article{cheng2024tv,
title={Open-TeleVision: Teleoperation with Immersive Active Visual Feedback},
author={Cheng, Xuxin and Li, Jialong and Yang, Shiqi and Yang, Ge and Wang, Xiaolong},
journal={arXiv preprint arXiv:2407.01512},
year={2024}
}

@article{levine2016end,
  title={End-to-end training of deep visuomotor policies},
  author={Levine, Sergey and Finn, Chelsea and Darrell, Trevor and Abbeel, Pieter},
  journal={Journal of Machine Learning Research},
  volume={17},
  number={39},
  year={2016}
}

@article{liu2025factr,
  title={FACTR: Force-Attending Curriculum Training for Contact-Rich Policy Learning}, 
  author={Jason Jingzhou Liu and Yulong Li and Kenneth Shaw and Tony Tao and Ruslan Salakhutdinov and Deepak Pathak},
  journal={Robotics: Science and Systems},
  year={2025}, 
}

@INPROCEEDINGS{Gello,
  author={Wu, Philipp and Shentu, Yide and Yi, Zhongke and Lin, Xingyu and Abbeel, Pieter},
  booktitle={IEEE/RSJ International Conference on Intelligent Robots and Systems (IROS)}, 
  title={GELLO: A General, Low-Cost, and Intuitive Teleoperation Framework for Robot Manipulators}, 
  year={2024},
}

@article{honerkamp2025whole,
  title={Whole-body teleoperation for mobile manipulation at zero added cost},
  author={Honerkamp, Daniel and Mahesheka, Harsh and von Hartz, Jan Ole and Welschehold, Tim and Valada, Abhinav},
  journal={IEEE Robotics and Automation Letters},
  year={2025},
  publisher={IEEE}
}

@article{jiang2025behavior,
  title={Behavior robot suite: Streamlining real-world whole-body manipulation for everyday household activities},
  author={Jiang, Yunfan and Zhang, Ruohan and Wong, Josiah and Wang, Chen and Ze, Yanjie and Yin, Hang and Gokmen, Cem and Song, Shuran and Wu, Jiajun and Fei-Fei, Li},
  journal={arXiv preprint arXiv:2503.05652},
  year={2025}
}

@inproceedings{ding2025bunny,
  title={Bunny-visionpro: Real-time bimanual dexterous teleoperation for imitation learning},
  author={Ding, Runyu and Qin, Yuzhe and Zhu, Jiyue and Jia, Chengzhe and Yang, Shiqi and Yang, Ruihan and Qi, Xiaojuan and Wang, Xiaolong},
  booktitle={2025 IEEE/RSJ International Conference on Intelligent Robots and Systems (IROS)},
  pages={12248--12255},
  year={2025},
  organization={IEEE}
}

@article{zhu2026emma,
  title={Emma: Scaling mobile manipulation via egocentric human data},
  author={Zhu, Lawrence Y and Kuppili, Pranav and Punamiya, Ryan and Aphiwetsa, Patcharapong and Patel, Dhruv and Kareer, Simar and Ha, Sehoon and Xu, Danfei},
  journal={IEEE Robotics and Automation Letters},
  year={2026},
  publisher={IEEE}
}

@article{intelligence2025pi_,
  title={$\pi_{0.5}$: a Vision-Language-Action Model with Open-World Generalization},
  author={Intelligence, Physical and Black, Kevin and Brown, Noah and Darpinian, James and Dhabalia, Karan and Driess, Danny and Esmail, Adnan and Equi, Michael and Finn, Chelsea and Fusai, Niccolo and others},
  journal={arXiv preprint arXiv:2504.16054},
  year={2025}
}
